\documentclass[runningheads]{llncs}
\usepackage{graphicx}
\usepackage{color}

\begin{document}
\title{Famous Companies Use More Letters in Logo:\\ A Large-Scale Analysis of Text Area in Logo}
\titlerunning{A Large-Scale Analysis of Text Areas in Logo}

%\author{Anonymous}
\author{Shintaro Nishi\and
Takeaki Kadota\and
Seiichi Uchida\orcidID{0000-0001-8592-7566}}
%]
%\authorrunning{S. Nishi et al.}
\institute{\vspace{-2mm}Kyushu University, Fukuoka, Japan}
\maketitle              % typeset the header of the contribution
\vspace{-0.8cm}
\begin{abstract}
%The abstract should briefly summarize the contents of the paper in
%15--250 words.
This paper analyzes a large number of logo images from the LLD-logo dataset, by recent deep learning-based techniques, to understand not only design trends of logo images and but also the correlation to their owner company. Especially, we focus on three correlations between logo images and their text areas, between the text areas and the number of followers on Twitter, and between the logo images and the number of followers. Various findings include the weak positive correlation between the text area ratio and the number of followers of the company. In addition, deep regression and deep ranking methods can catch correlations between the logo images and the number of followers.

\end{abstract}
%
%
%%%%%%%%%%%%%%%%%%%%%%%%%%%%%%%%%%%%%%%%%%%%%%%%%%%%%%%%%%%
\section{Introduction} \label{sec:intro}
%%%%%%%%%%%%%%%%%%%%%%%%%%%%%%%%%%%%%%%%%%%%%%%%%%%%%%%%%%%
%ロゴは大事．企業の印象を決定づける．
Logo is a graphic design for the public identification of a company or an organization. Therefore, 
each logo is carefully created by a professional designer, while considering various aspects of 
the company. In other words, each logo represents the policy, history, philosophy, commercial strategy, etc., of the company, along with its visual publicity aim.\par
\begin{figure}[t]
 \begin{minipage}{0.30\textwidth}
  \centering
   \includegraphics[width=\textwidth]{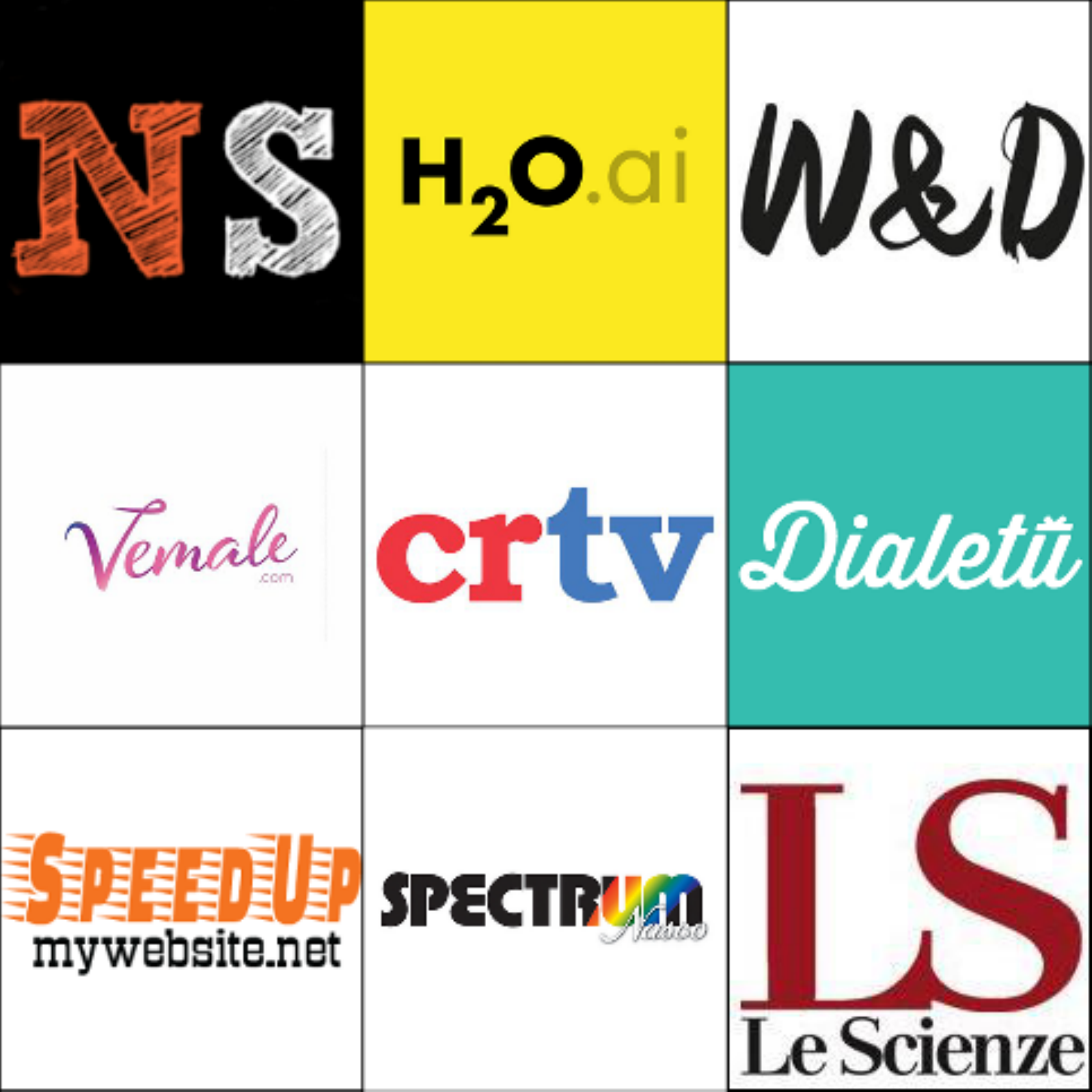}\\
      (a)~Logotype.
 \end{minipage}
 \ 
 \begin{minipage}{0.30\textwidth}
  \centering
  \includegraphics[width=\textwidth]{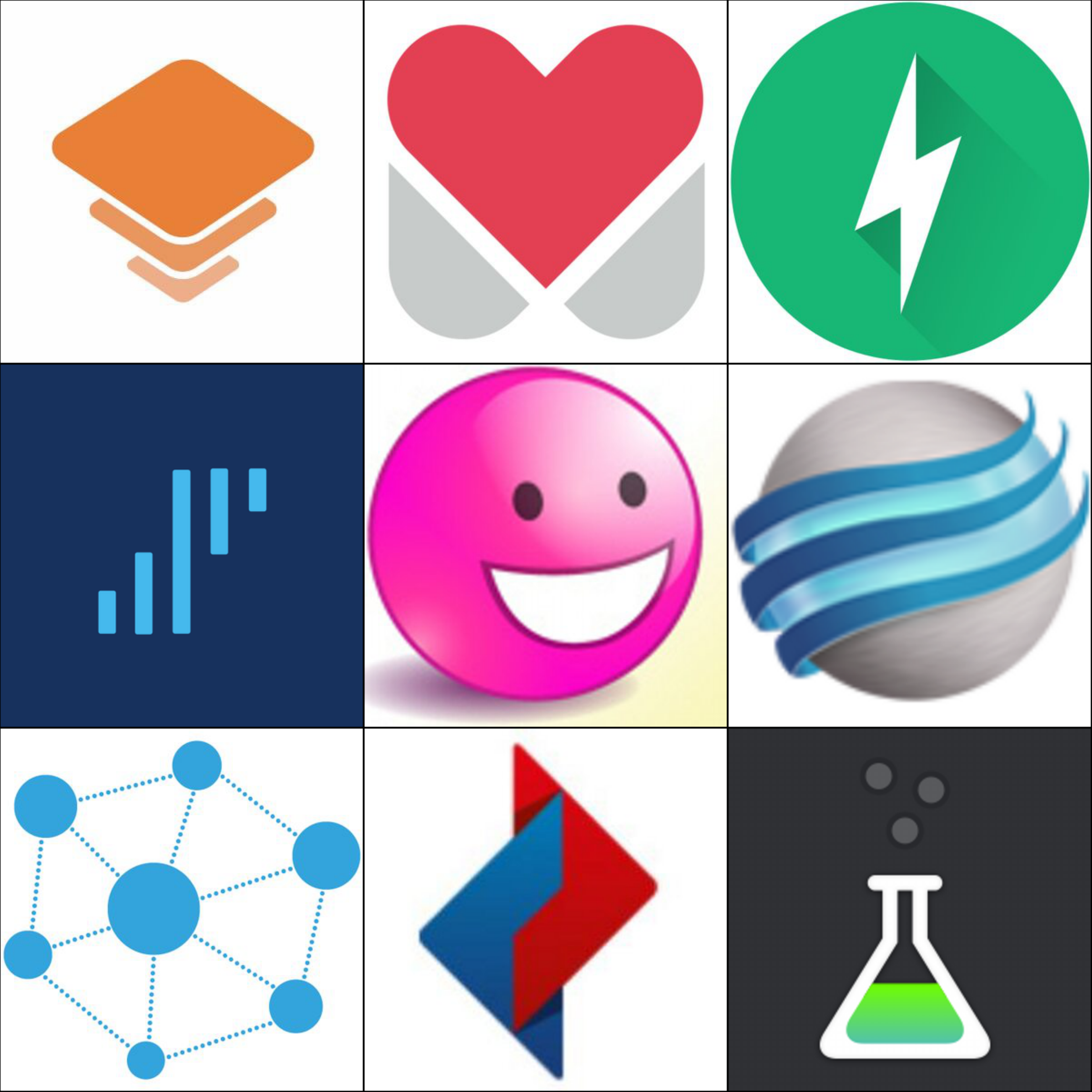}\\
   (b)~Logo symbol.
 \end{minipage}
 \ 
 \begin{minipage}{0.30\textwidth}
  \centering
  \includegraphics[width=\textwidth]{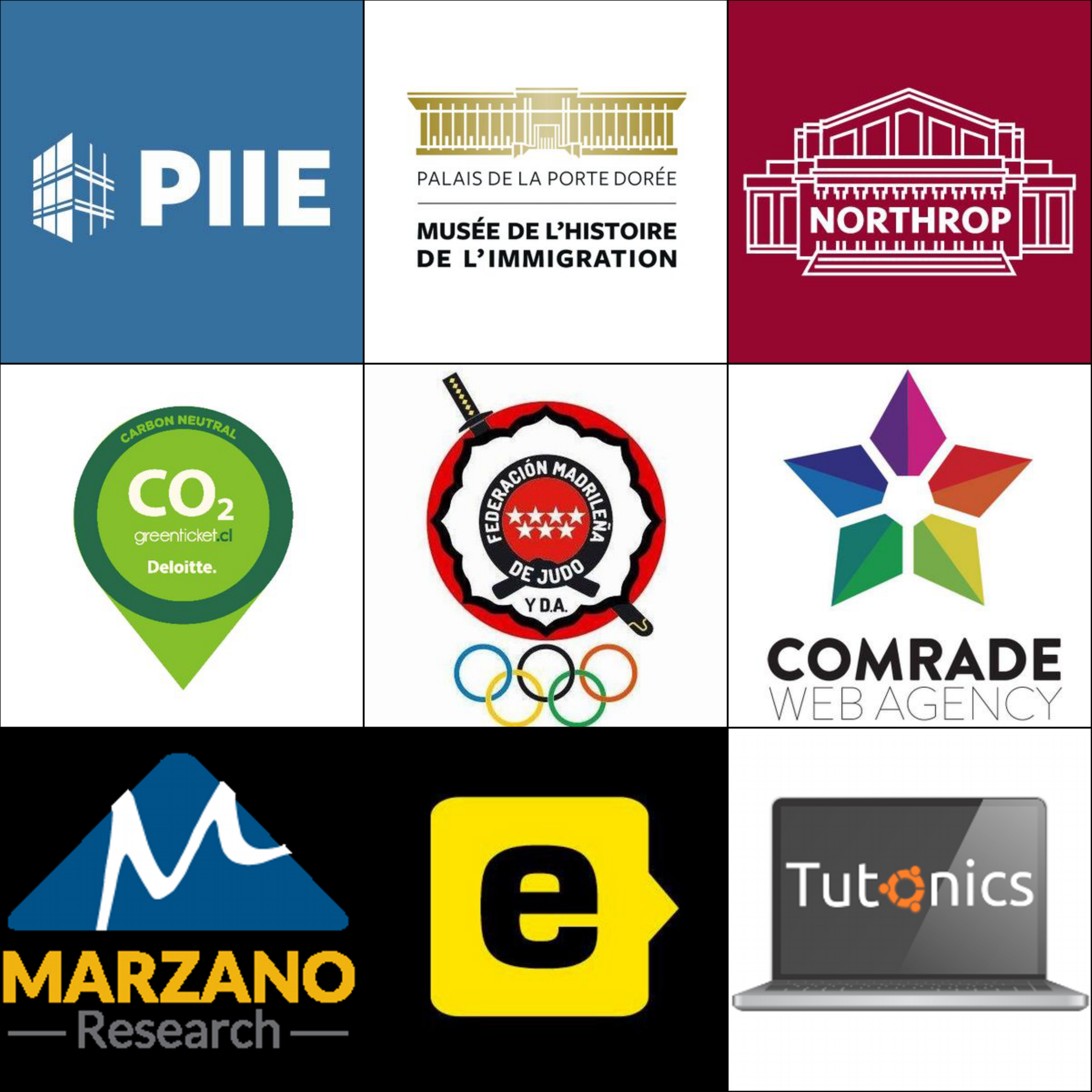}\\
   (c)~Mixed.
 \end{minipage}\\[-3mm]
 \caption{Logo image examples.} %\color{red}The attached numbers show the number of followers of the corresponding company on Twitter.\color{black}
 \label{fig:logo-example}
\end{figure}

%
%ロゴの種類：テキストのみ，マークのみ，両方
Logo can be classified roughly into three types: logotype, logo symbol, and their mixed~\footnote{There are several variations in the logo type classification. For example, in \cite{Adir:2012}, three types are called ``text-based logo'', ``iconic or symbolic logo'', and ``mixed logo'', respectively.}. Fig.~\ref{fig:logo-example} shows several examples of each type. {\em Logotype} is a logo comprised of only letters and often shows a company name or its initial letters. {\em Logo symbol} is a logo comprised of some abstract mark or icon or pictogram. {\em Mixed} is a logo comprised of a mixture of logotype and logo symbol.\par

% 本論文の全体的な目的はロゴ分析である．そうした分析を行うことの価値
The general purpose of this paper is to analyze the logo images from various aspects to understand the relationship among their visual appearance, text area, and the popularity of the company. For this purpose, we utilize various state-of-the-art techniques of text detection and image clustering, and a large logo dataset, called LLD-logo~\cite{Sage:2018}. 
To the authors' best knowledge, logo image analysis has not been attempted on a large scale or objective way using recent machine-learning-based image analysis techniques. Through the analysis, it is possible to reveal not only the trends of logo design but also the relationship between a company and its logo. The analysis results are helpful to design new logos that are appropriate to the company's intention.\par
%
% 本論文の第一の分析は，ロゴにおける文字の使用率．
The first analysis is how texts are used in logos. Specifically, we detect the text area in each logo image; if no text, the logo image is a logo symbol. If text only, it is a logotype. Otherwise, it is a mixed logo. Using 122,920 logo images from \cite{Sage:2018} and a recent text detection technique~\cite{baek2019character}, it is possible to understand the ratio of the three logo classes. Moreover, it is possible to quantify the text area ratio and the text location in mixed logos.\par
%
% 本論文の第二の分析は，ロゴ内のテキスト量と人気度の関係
The second and more important analysis is the relationship between the text area ratio and the popularity of the company. The logo images in the LLD-logo dataset are collected from Twitter. Therefore it is possible to know {\em the number of followers}, which is a good measure of the company's popularity. Since most logotypes are company names, the analysis will give a hint to answering whether famous companies tend to appeal their names in their logo or not. Since this relationship is very subtle, we try to catch it by a coarse view using DeepCluster~\cite{Caron2018deepcluster}.
\par
% 本論文の第三の分析は，DeepClusterRankerを用いたロゴデザインと人気度の関係解明．
The last analysis is the relationship between the whole logo image (including both text areas and symbols) and the company's popularity by using regression analysis and ranking analysis. If it is possible to realize a regression function and/or a ranking function with reasonably high accuracy, they are very useful as references of better logo design that fits the company's popularity. 
%However, the relationship between the logo image and the popularity will be very weak because companies with similar popularity will {\em not} use logos with similar appearances. In order to catch the weak 
%relationship, DeepClusterRanker performs logo image clustering and cluster ranking simultaneously in a deep neural network-based framework. As the result, visually similar logos are gathered into a cluster and all the clusters are ordered according to the average popularity of the cluster members. In other words, DeepClusterRanker will catch the weak relationship by the coarse (i.e., cluster-wise) ranking. \par
%
The main contributions of this paper are summarized as follows:
\begin{itemize}
    \item This paper provides the first large-scale objective analysis of the relationship among logo images, texts in logo, and company popularity. The analysis results will give various hints for the logo design strategy.
    \item The robust estimation using DeepCluster revealed the positive correlation between the popularity of a company and the text area ratio in its logo.
    \item Regression and ranking analyses showed the possibility of estimating the absolute popularity (i.e., the number of followers) or relative popularity from logo images; this result suggests we can evaluate the goodness of the logo design by the learned regression and ranking functions.  
    \item In addition to the above results, we derive several reliable statistics, such as the ratio among three logo types (logotypes, logo symbols, and mixed logos) and the text area ratio and text location in logo images. 
\end{itemize}

%%%%%%%%%%%%%%%%%%%%%%%%%%%%%%%%%%%%%%%%%%%%%%%%%%%%%%%%%%%
\section{Related Work}\label{sec:related}
%%%%%%%%%%%%%%%%%%%%%%%%%%%%%%%%%%%%%%%%%%%%%%%%%%%%%%%%%%%
\subsection{Logo design analysis}
Logo design has been studied from several aspects, especially, marketing research.
According to A\^idr et al.~\cite{Adir:2012}, logo is classified into three types, 
``a text defined a logo'', ``iconic or symbolic logo'', and ``A mixed logo.'' In this paper, we call them logotype, logo symbol, and mixed logo, respectively. A brief survey by Zao~\cite{Zhao2017} discusses the recent diversity of logos.
Sundar et al.~\cite{Sundar2014} analyzed the effect of the vertical position of a logo on the package. Luffarelli et al.~\cite{luffarelli2019visual} analyzed how the symmetry and asymmetry in logo design give different impressions by using hundreds of crowd-workers; they conclude that asymmetric logos give more {\it arousing} impressions. \par
Recently, Luffarelli et al.~\cite{luffarelli2019} report an inspiring result that the {\it descriptiveness} of logos positively affects the brand evaluation. Here, the descriptiveness means that the logo describes explicitly what the company is doing by texts and illustrations. This report, which is also based on subjective analysis using crowd-workers, gives insights into the importance of texts in logo design. \par
Computer science has recently started to deal with logo design, stimulated by public datasets of logo images. Earlier datasets, such as UMD-Logo-Database\footnote{Not available now. According to \cite{Neumann2002}, it contained 123 logo images.} were small. In contrast, nowadays, the dataset size becomes much larger.
WebLogo-2M~\cite{Su:2017} is the earliest large dataset prepared for the logo detection task. The dataset contains 194 different logos captured in 1,867,177 web images. Then, Sage et al.~\cite{Sage:2018} have published the LLD-logo dataset, along with the LLD-icons dataset. The former contains 122,920 logo images collected from Twitter
and the latter 486,377 favicon images (32$\times$ 32 pixels) via web-crawling. They used the dataset for logo synthesis. Recently, Logo-2k+ dataset is released by Wang et al.~\cite{wang2020logo}, which contains 2,341 different logos in 167,140 images.
The highlight of this dataset is that the logo images collected via web-crawling are classified into 10 company classes (Food, Clothes, Institution, Accessories, etc.). In ~\cite{wang2020logo}, the logo images are applied to a logo classification task.\par
Research to quantify logo designs by some objective methodologies is still not so 
common. One exceptional trial is Karamatsu et al.~\cite{Karamatsu2019LogoDA}, where the favicons from LLD-icons are analyzed by a top-rank learning method for understanding the trends of favicon designs in each company type. This paper also attempts to quantify the logo design by analyzing the relationships among logo images, text areas in them, and their popularity (given as the number of followers on Twitter).

%-----------------------------------------------
\subsection{Clustering and ranking with deep representation}
We will use DeepCluster~\cite{Caron2018deepcluster} for clustering the LLD-logo images. Recently, the combination of clustering techniques with deep representation becomes popular~\cite{Aljalbout2018,Min2018}. DeepCluster is a so-called self-supervised technique for clustering, i.e., an unsupervised learning task, and has already been extended to tackle new tasks. For example, Zhan et al.~\cite{Zhan_2020_CVPR}
developed an online deep clustering method to avoid the alternative training procedure in DeepCluster. Tang et al.~\cite{Tang2020} apply deep clustering to unsupervised domain adaptation tasks.\par
In this paper, we combine DeepCluster with the learning-to-rank technique. Specifically, we will combine RankNet~\cite{Burges2005} (originally with multi-layer perceptron) with the convolutional neural network trained as DeepCluster. The idea of RankNet is also applied to many applications, such as image attractiveness evaluation~\cite{Ma2019}.
The recent development from RankNet is well-summarized \cite{Guo2020}.
%
%%%%%%%%%%%%%%%%%%%%%%%%%%%%%%%%%%%%%%%%%%%%%%%%%%%%%%%%%%%
\section{Logo image dataset --- LLD-logo\cite{Sage:2018}\label{sec:data}}
%%%%%%%%%%%%%%%%%%%%%%%%%%%%%%%%%%%%%%%%%%%%%%%%%%%%%%%%%%%
In this paper, we use the LLD-logo dataset by Sage et al.~\cite{Sage:2018}. LLD-logo contains 122,920 logo images collected from Twitter profile images using its API called {\tt Tweepy}. Several careful filtering operations have been made on the collection to exclude facial images, harmful images and illustrations, etc. 
The size of the individual logo images in the dataset is $63\times63 \sim 400\times 400$.
\par
The logo images in the LLD-logo dataset have two advantages over the logo image datasets that contain camera-captured images, such as Logo-2K~\cite{wang2020logo}. First, logo images from LLD-logo are so-called born-digital and thus have no disturbance from uneven light, geometric distortion, low resolution, blur, etc.\par
\begin{figure}[t]
    \centering
    \includegraphics[width=0.8\linewidth]{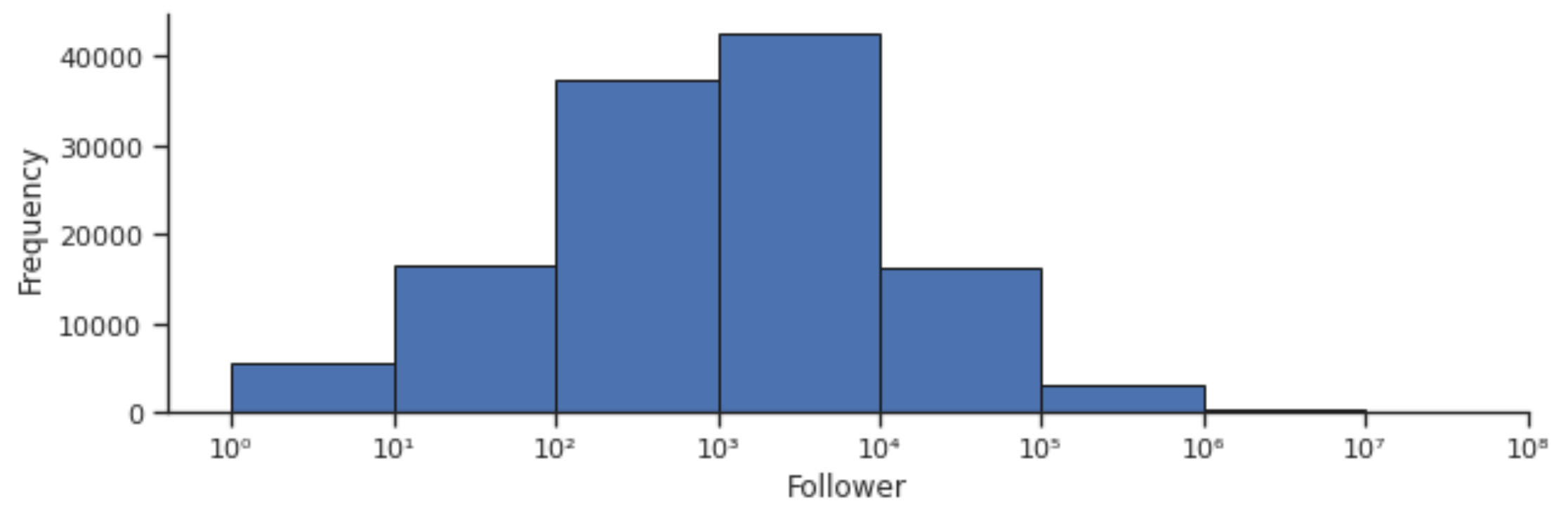}\\[-5mm]
    \caption{Distribution of followers of the companies listed in LLD-logo dataset.}
    \label{fig:popu_dist}
\end{figure}

The second advantage of LLD-logo is meta-data, including {\it the number of followers}, which is a very good index to understand the popularity of the individual companies\footnote{The number of followers is provided in the file {\tt LLD-logo.hdf5}, which is provided with LLD-logo image data. Precisely, the resource of {\tt meta\_data/twitter/user\_objects} in this hdf5 file contains the {\tt followers\_count} data, which corresponds to the number of followers.}. Fig.~\ref{fig:popu_dist} shows the distribution of the number of followers of the companies in the LLD-logo dataset. Note that the bins of this histogram are logarithmic; this is a common way to analyze the popularity, especially the Twitter follower distributions~\cite{ardon2011spatio,bakshy2011everyone,lerman2012social,stringhini2013follow}. This distribution shows that the companies with $10^2\sim10^4$ followers are the majority and there are several exceptional companies with less than $10$ followers or more than $10^7$ followers\footnote{For the logarithmic plot, we exclude the companies with zero follower. The number of such companies is around 1,000 and thus no serious effect in our discussion.}.
\par
%% 人気度のディストリビューションもここで．

\begin{figure}[t]
 \begin{minipage}{0.30\textwidth}
  \centering
   \includegraphics[width=\textwidth]{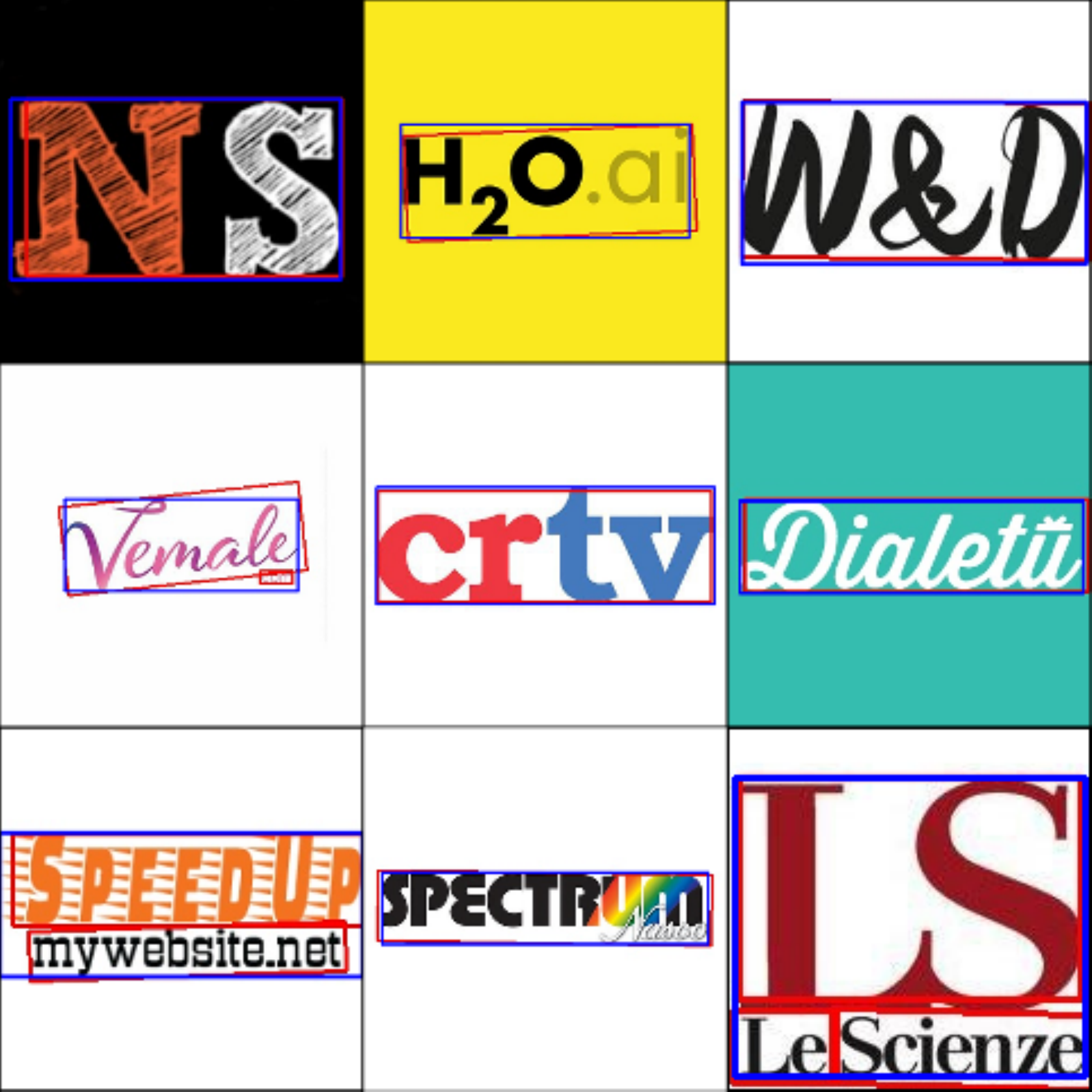}\\
      (a)~Logotype.
 \end{minipage}
 \ 
 \begin{minipage}{0.30\textwidth}
  \centering
  \includegraphics[width=\textwidth]{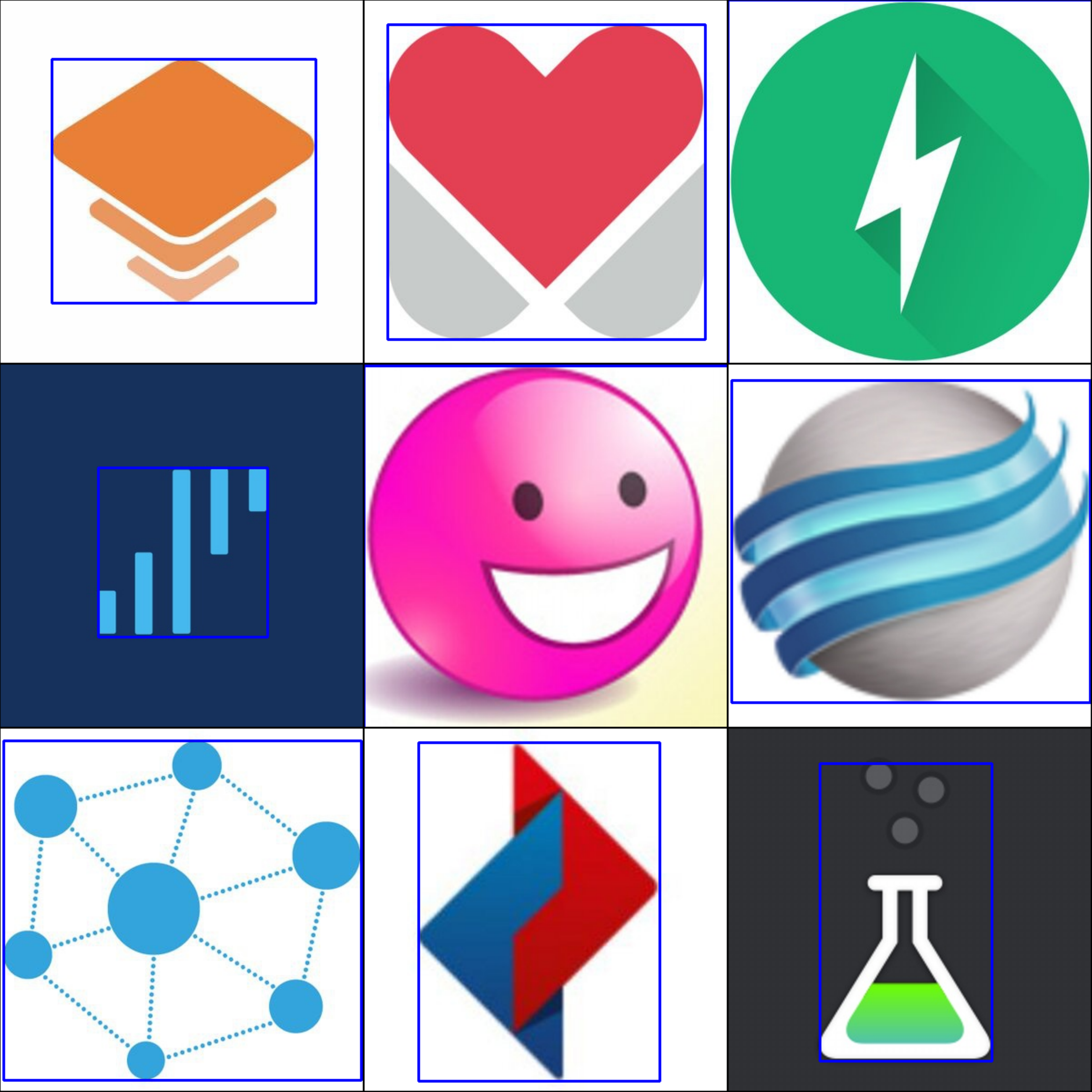}\\
   (b)~Logo symbol.
 \end{minipage}
 \ 
 \begin{minipage}{0.30\textwidth}
  \centering
  \includegraphics[width=\textwidth]{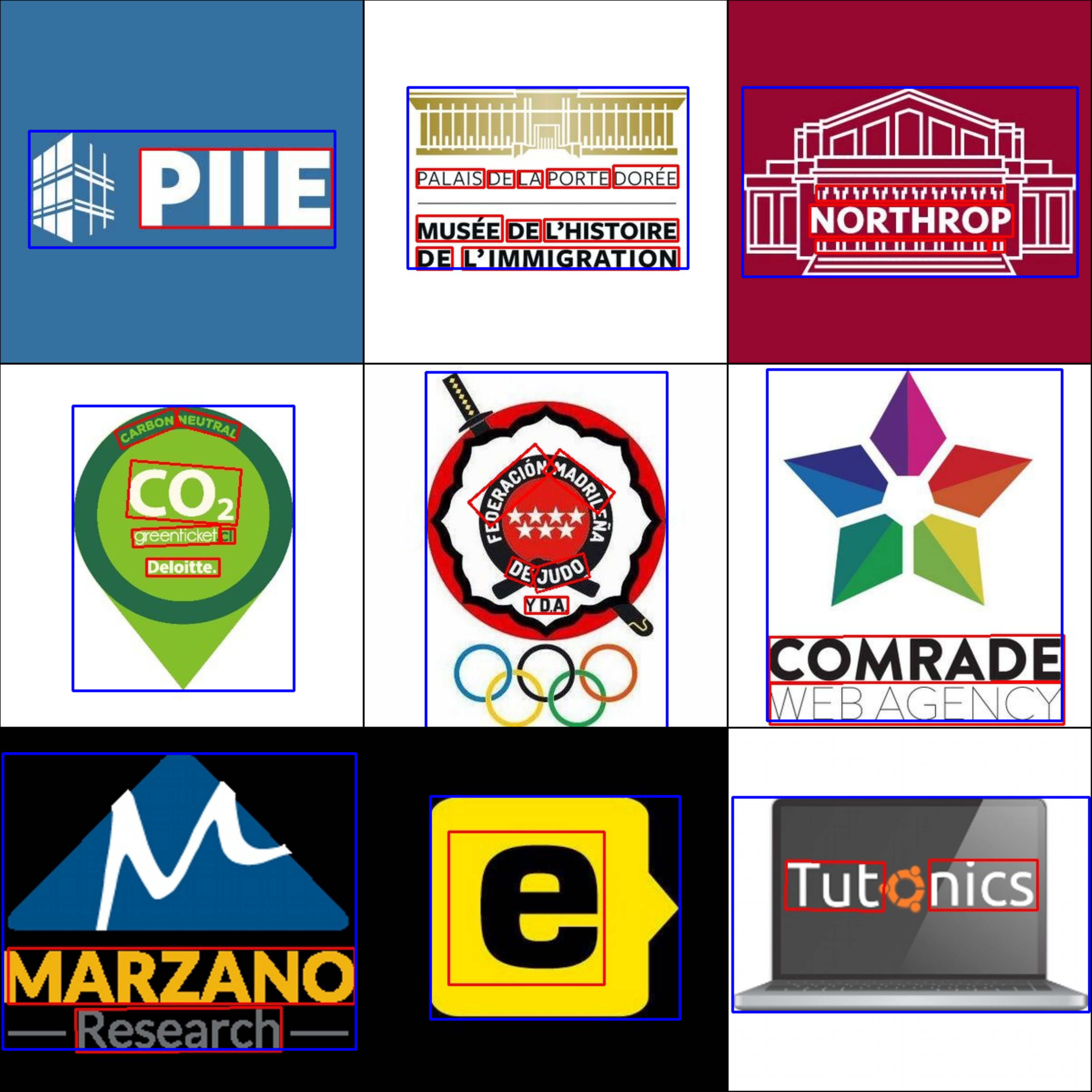}\\
   (c)~Mixed.
 \end{minipage}
%    \centering
%    \includegraphics[width=\linewidth]{images/CRAFT_result.pdf}\\[-3mm]
    \caption{Text detection from logo images by CRAFT~\cite{baek2019character}. The red boxes are the detected text areas, whereas the blue boxes are the bounding box of the whole logo image.}
    \label{fig:CRAFT}
\end{figure}
%%%%%%%%%%%%%%%%%%%%%%%%%%%%%%%%%%%%%%%%%%%%%%%%%%%%%%%%%%%
\section{Analysis 1: How Much Are Texts Used in Logo?}
%%%%%%%%%%%%%%%%%%%%%%%%%%%%%%%%%%%%%%%%%%%%%%%%%%%%%%%%%%%
\subsection{Text detection in logo image}
As shown in Fig.~\ref{fig:logo-example}, logo images often include texts. Especially, a logotype is just a text (including a single letter); that is, the text fills 100\% of the logo area. In contrast, a logo symbol does not include any text. A mixed log will have a different text area ratio. In other words, by observing the text area ratio in the individual logo images, we can understand not only the frequency of the three types, but also the ratio trend in the mixed type.\par
Recent scene text detection techniques allow us to extract the text area from a logo image, even when the texts are heavily decorated, rotated, and overlaid on a complex background. We use CRAFT~\cite{baek2019character}, which is one of the state-of-the-art scene text detectors. Fig.~\ref{fig:CRAFT} shows several text detection results by CRAFT. They prove that CRAFT can detect the texts in logos very accurately, even with the above difficulties. We, therefore, will use the text bounding boxes detected by CRAFT in the following experiments.\par
Note that the detection results by CRAFT are not completely perfect. Especially, in logo designs, the border between texts and illustrations is very ambiguous. For example, the logo of ``MARZANO'', the calligraphic `M' in the blue triangle is missed. The logo of ``Tutonics'', `o' is not detected as a letter due to its illustration-like design. As indicated by those examples, it is impossible to expect to have perfect extraction results from the logo images. Even so, CRAFT still extracts truly text parts quite accurately in most cases.

%------------------------------------------------------
\subsection{Text area ratio and the number of text boxes}
In the following analysis, we often use two simple metrics: text area ratio and the number of text boxes. The text area ratio is the ratio of the text area in the whole logo area. The text area is specified by all the bounding boxes (red boxes in Fig.~\ref{fig:CRAFT}) in the logo image. The whole logo area, which is shown as the blue bounding box in Fig.~\ref{fig:CRAFT}, 
is defined by the bounding box of the design element on the logo. Therefore, the whole logo area is often smaller than the image size. For example, the whole logo area of the logo of ``NS'' 
is almost the same as the text bounding box, and thus its text area ratio is about 100\%\footnote{As we will see later, the text area ratio sometimes exceeds 100\%. There are several reasons behind it; one major reason is the ambiguity of the whole logo's bounding box and the individual text areas' bounding boxes. If the latter slightly becomes the former, the ratio exceeds 100\%. Some elaborated post-processing might reduce those cases; however, they do not significantly affect our median-based analysis.}.\par
The number of text boxes is simply defined as the number of text boxes detected by CRAFT.  Since CRAFT can separate a multi-line text into the lines and then give the boxes at each line, 
the number of text boxes increases with the number of lines. If there is sufficient space between two words in a text line, CRAFT gives different text boxes for the words. If not, CRAFT will give a single box, as shown in the ``SpeedUp'' logo.

%------------------------------------------------------
\subsection{The ratio of three logo types}
As the first analysis, we classify each logo image into one of three logo types using the text area ratio. This classification is straightforward: if the text area ratio is 0\%, it will be a logo symbol. If the text area ratio is more than 90\%, we treat it as a logotype. Otherwise, it is a mixed logo. Note that we set 90\% as the class boundary between logotype and mixed (instead of 100\%) because the bounding boxes of the whole logo and the text area are not always completely the same, even for logotypes.\par
The classification result  of all  122,920 logos in the large logo set shows that the ratio of three types are: 4\%(logotype) : 26\%(logo symbol) : 70\% (mixed). The fact that mixed logos are clear majority indicates that most companies show their name with some symbol in their logo. Surprisingly, logotypes are very minority. Logo symbols are much more than logotypes but less than half of mixed logos.

\begin{figure}[t]
\centering
\begin{minipage}[b]{0.57\textwidth}
    \centering
    \includegraphics[width=\linewidth]{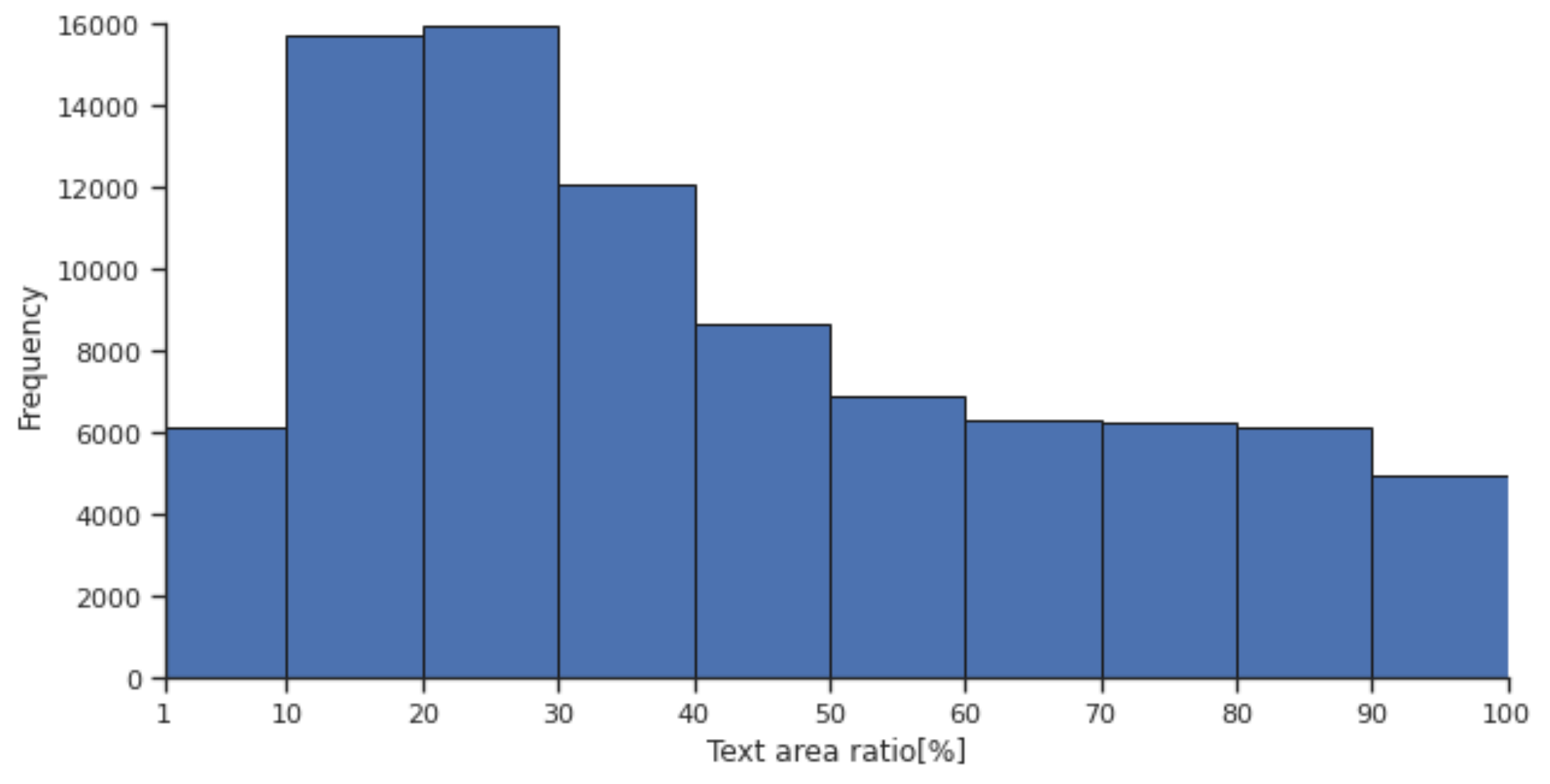}\\[-1mm]
    (a)
\end{minipage}
\begin{minipage}[b]{0.4\textwidth}
    \centering
    \includegraphics[width=\linewidth]{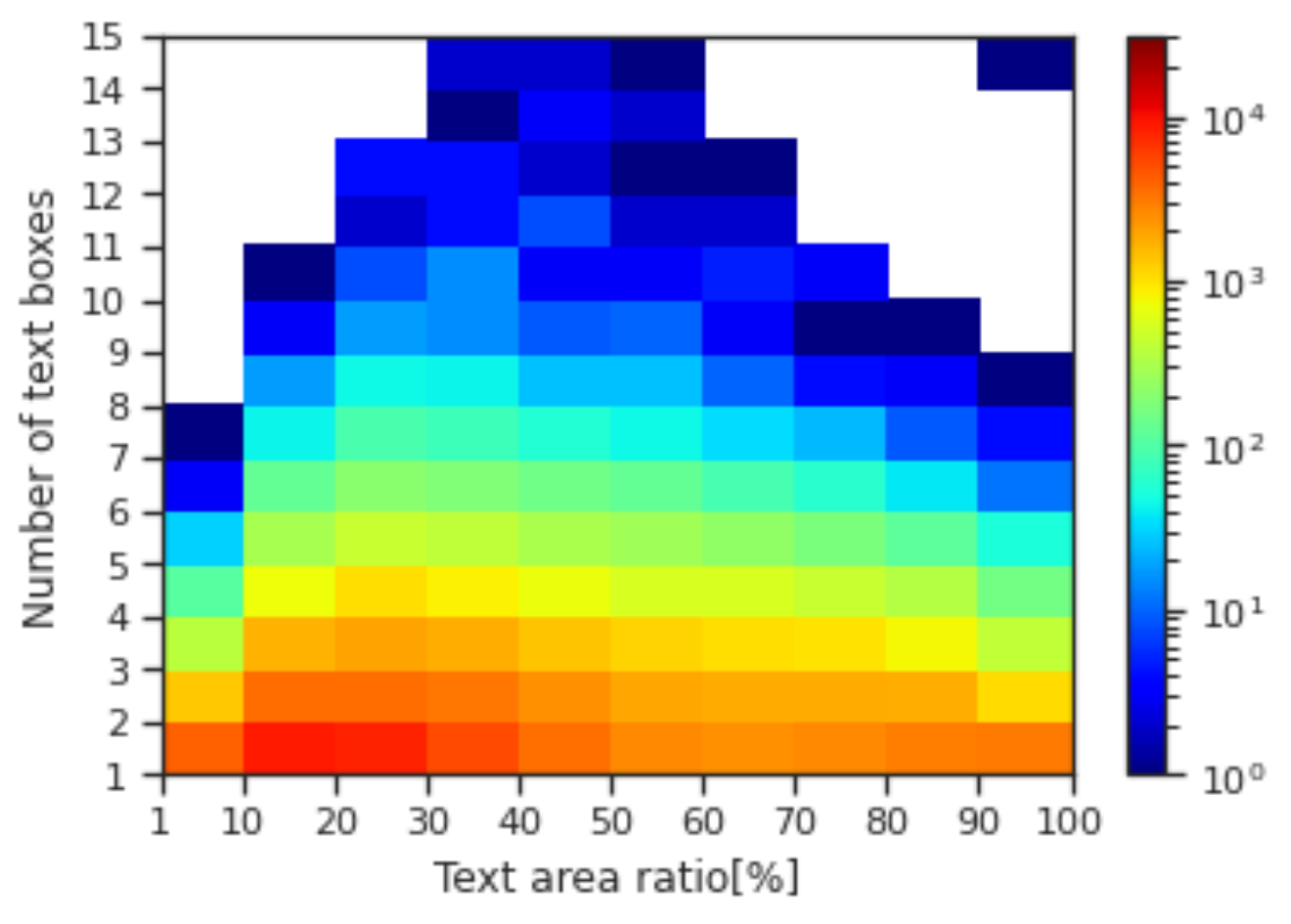}\\[-1mm]
 (b)    
\end{minipage}\vspace{-3mm}
    \caption{(a)~Distribution of the text area ratios. (b)Two-dimensional histogram showing the relationship between the text area ratio and the number of text boxes.}
    \label{fig:dist-text-area-ratios}
\medskip
    \includegraphics[width=0.6\linewidth]{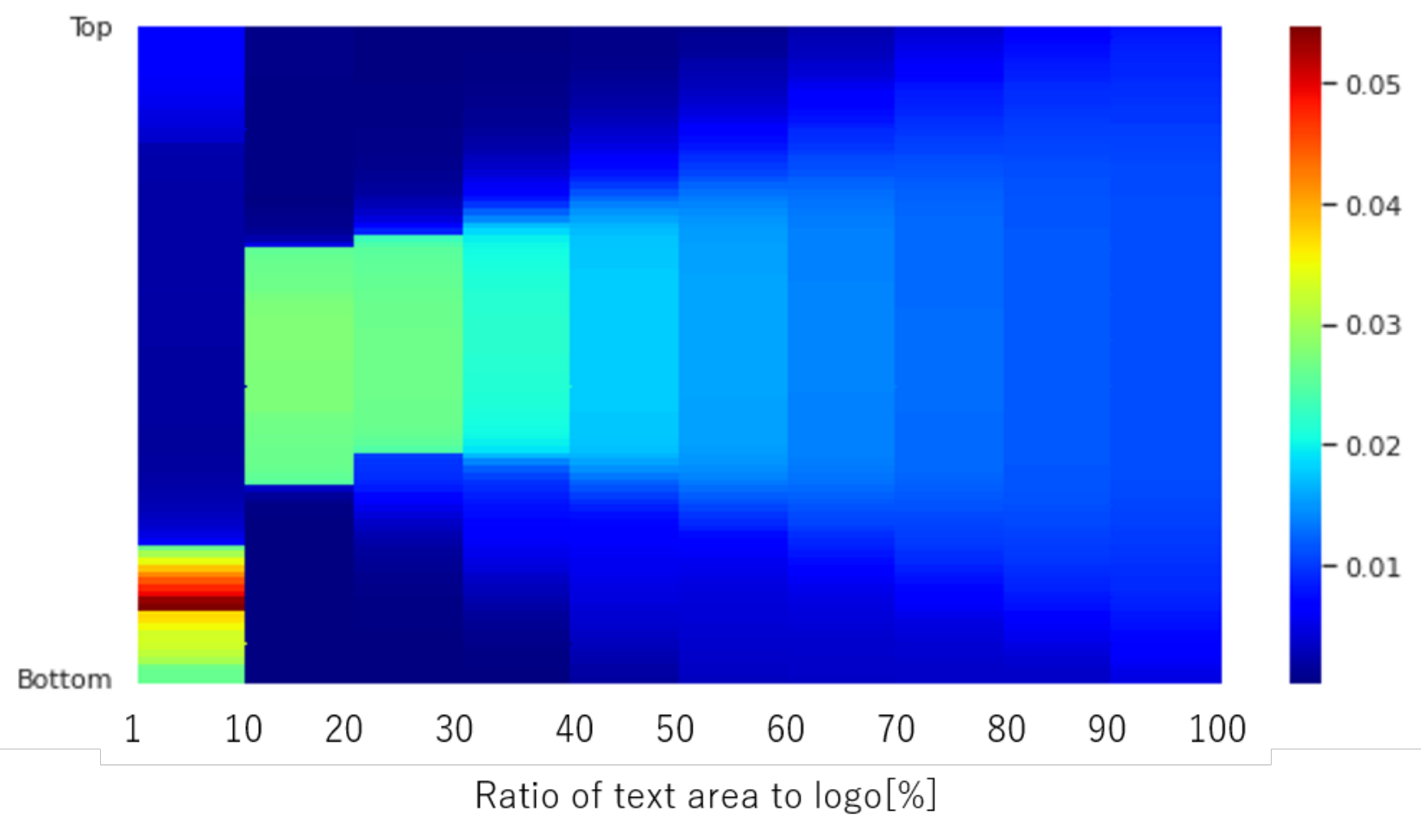}\\[-5mm]
    \caption{Distribution of text regions in the vertical direction for different text area ratios. The frequency is normalized at each vertical bin (i.e., each text area ratio). }
    \label{fig:vertical-location}
\medskip
    \includegraphics[width=0.8\linewidth]{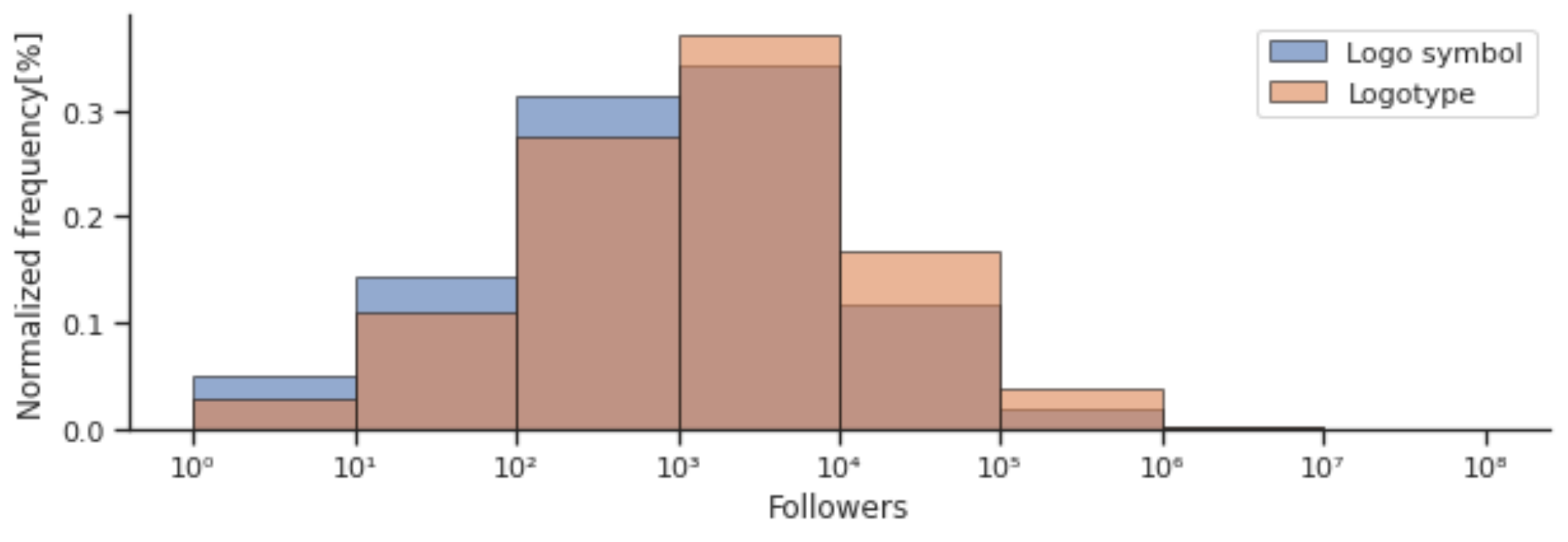}\\[-5mm]
    \caption{Distributions of Twitter followers for logotype and logo symbol.}
    \vspace{-3mm}
    \label{fig:follower-compare-1}

\end{figure}
%------------------------------------------------------
\subsection{Distribution of the text area ratio}
Fig.~\ref{fig:dist-text-area-ratios}~(a) shows the histogram of the text area ratios, where logo symbols are excluded. This suggests that the text areas often occupy only $10\sim 30\%$ of the whole logo area. Fig.~\ref{fig:dist-text-area-ratios}~(b) shows the two-dimensional histogram to understand the relationship between the text area ratio and the number of text boxes. The histogram says that the most frequent case is a single text box with $10\sim 30\%$ occupancy. Logos with multiple text boxes exist but not so many. It should also be noted that the text area ratio and the number of text boxes are not positively correlated; even if the ratio increases, the text boxes do not.

%------------------------------------------------------
\subsection{The location of the text area}
Fig.~\ref{fig:vertical-location} visualizes the vertical location of the text area for different text area ratios. Since the logo height is normalized for this visualization, the top and bottom of the histogram correspond to the top and bottom of logo images, respectively. This visualization reveals that when a small text area ($<10\%$) is presented on a logo image, it is mostly located at the bottom of the logo. However, when the text area occupies more than $10\%$ of the whole logo area, it is mostly located around the center of the logo.
%------------------------------------------------------
\subsection{Does a famous company show the name on its logo or not?\label{sec:two-hists}}
As noted in Section~\ref{sec:data}, the LLD-logo dataset includes the number of followers for each logo (i.e., company). Fig.~\ref{fig:follower-compare-1} shows the distributions of the follower
numbers for two extreme logo types, i.e., logotype and logo symbol. For each type, the distribution is normalized so that its total becomes 1. As noted before, the number of followers is treated in a logarithmic manner according to the standard way. \par
These distributions prove that the average number of followers is larger when the logo is logotypes than logo symbols. One might have supposed that popular companies need not show their name (because they are always famous enough) and use logo symbols more. However, among the logotype users, the famous companies are not minority~\footnote{Recall that each distribution is normalized. As noted before, logotypes are just 4\%, whereas logo symbols are 26\%. Thus, as the absolute numbers, the famous companies use more logo symbols than logotypes.}.  The relationship between the number of followers and the text area ratio will be analyzed in more detail in Section~\ref{sec:correlation}.

%%%%%%%%%%%%%%%%%%%%%%%%%%%%%%%%%%%%%%%%%%%%%%%%%%%%%%%%%%%
\section{Analysis 2: Cluster-Wise Correlation Analysis between the Number of Followers and the Text Area Ratio}
%
%
%%%%%%%%%%%%%%%%%%%%%%%%%%%%%%%%%%%%%%%%%%%%%%%%%%%%%%%%%%%
%DeepClusterによるロゴクラスタリング
% DeepClusterの手法
% クラスタサイズ(a)
% フォロワー数のメディアンで降順に並べた箱ひげ図(b)
% クラスタごとのtext area ratioの箱ひげ＋「メディアン」の「メディアンフィルタ平滑化」(c)
% 上のグラフで注目したいくつかのクラスタを見る（画像例）
% - ゴミクラスタ（巨大クラスタ）
% - 文字がやたら多いクラスタ
% - 文字がほとんどないクラスタ
% - 文字量の分散が大きいクラスタ，小さいクラスタ
%         ↑各画像例10個ぐらい
%%%%%%%%%%%%%%%%%%%%%%%%%%%%%%%%%%%%%%%%%%%%%%%%%%%%%%%%%%%
\begin{figure}[t]
    \centering
    \begin{minipage}{0.03\textwidth}(a)\end{minipage}
    \begin{minipage}{0.95\textwidth}\hfill
    \includegraphics[width=\linewidth]{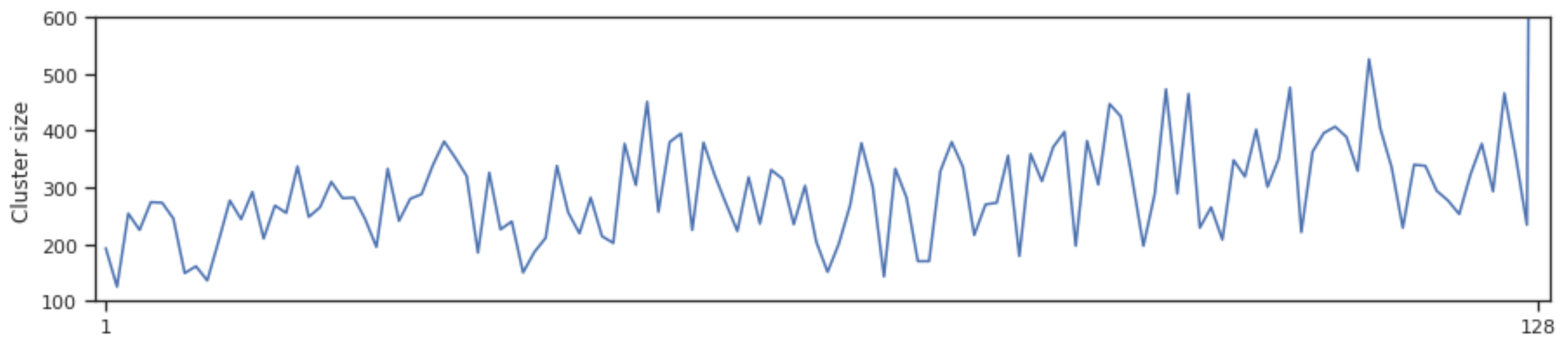}
    \end{minipage}\\[-1mm]
    \begin{minipage}{0.03\textwidth} (b)\end{minipage}
    \begin{minipage}{0.95\textwidth}\hfill
    \includegraphics[width=\linewidth]{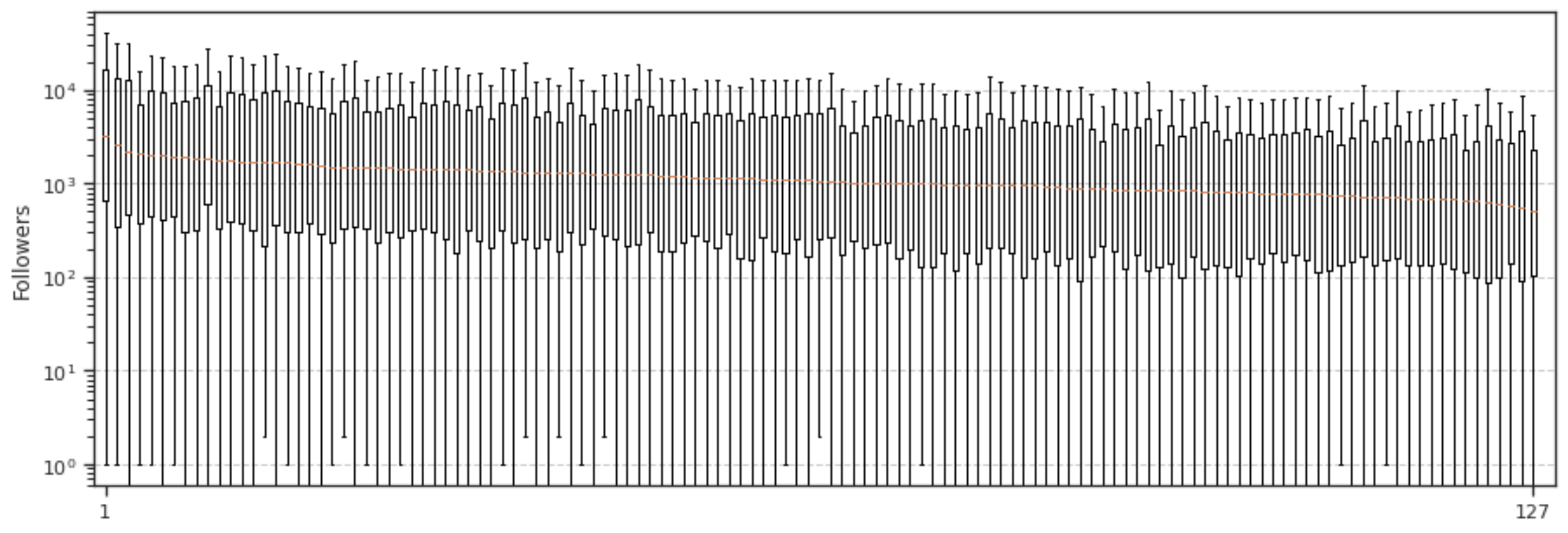}
    \end{minipage}\\[-1mm]
    \begin{minipage}{0.03\textwidth}(c)\end{minipage}
    \begin{minipage}{0.95\textwidth}\hfill
    \includegraphics[width=\linewidth]{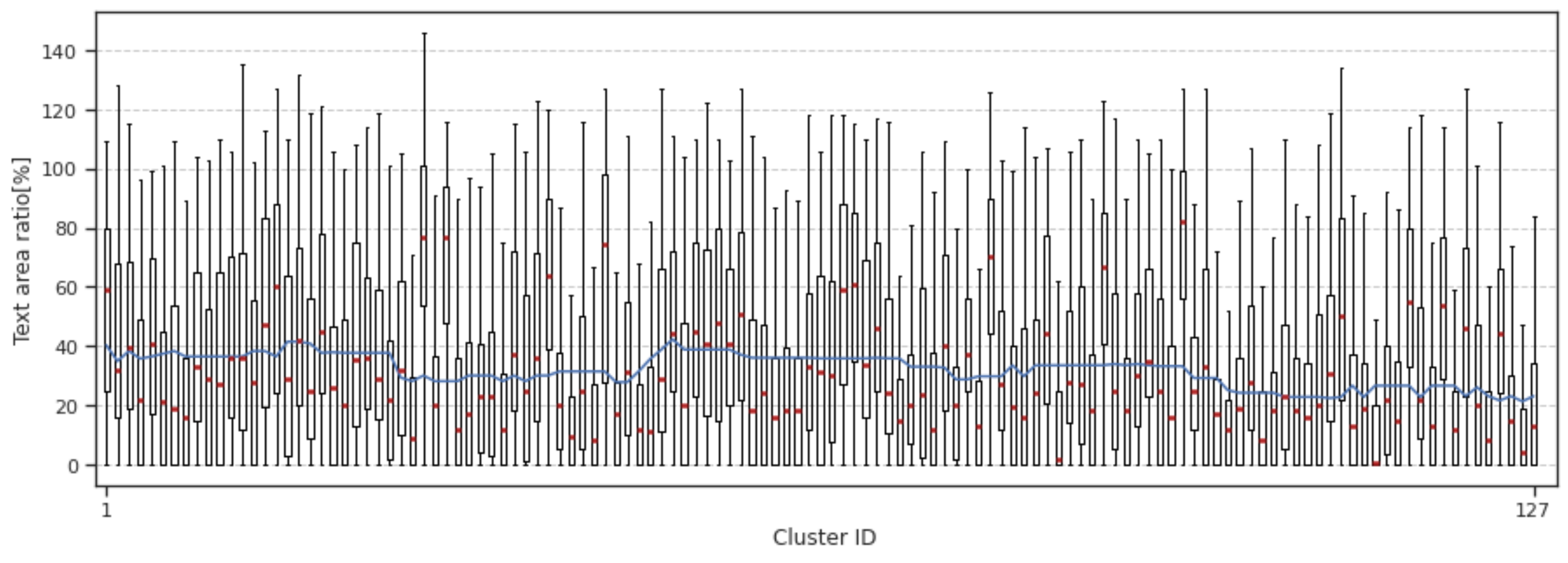}
    \end{minipage}\\[-1mm]
    \begin{minipage}{0.03\textwidth}(d)\end{minipage}
    \begin{minipage}{0.95\textwidth}\hfill
    \includegraphics[width=0.97\linewidth]{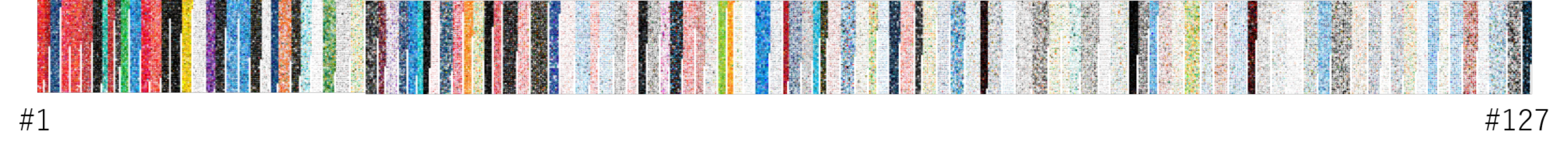}\\[-3mm]
    \end{minipage}\\[-1mm]    
  
    \caption{(a)~Cluster size $s_i$, (b)~the box plot of the numbers of followers, and (c)~the box plot of text area ratios, for each cluster $i$. In (a)-(c), the horizontal axis is the cluster ID $i$ and sorted by the descending order of $f_i$ (orange), which is the median value of the number of followers. The blue curve in (c) shows a smoothed transition of $t_i$ (red), which is the median values of text area ratios. (d)~Overview of all clusters. From left to right, the logo images of each cluster are shown as very tiny images.}
    \label{fig:cluster-wise-area-distribution.}
\end{figure}
%------------------------------------------------------
\subsection{Logo image clustering by DeepCluster~\cite{Caron2018deepcluster}}
DeepCluster is a clustering technique with a representation learning function. Its idea is rather simple based on the typical pseudo-labeling technique. Given a set of image samples, DeepCluster first outputs their representations (i.e., feature vectors) via convolutional layers. Second, the $k$-means clustering is performed to the feature vectors, after applying PCA for dimensionality reduction. Third, the cluster ID of each pseudo-label is assigned to each sample as its pseudo-label. Fourth, the convolutional layers are re-trained along with the attached classification layers so that the whole neural network can classify the samples according to their pseudo-labels. By repeating those steps, we have $k$ clusters for the set. Since DeepSets relies on a self-trained discriminability, we do not need to try various similarity metrics for the suitable clustering result. Moreover, we can expect that it can capture the similarity among logos in rather an abstract level (than the bitmap level), based on the fact that it shows accurate performance at the classification, detection, and segmentation task on Pascal VOC~\cite{Caron2018deepcluster}.\par
We apply DeepCluster to 40,000 logo images randomly selected from LLD-logo. 
The number of clusters is fixed at $k=128$. All the images are rescaled to be 
$224\times 224$ pixels. Fig.~\ref{fig:cluster-wise-area-distribution.}~(a) shows the size $s_i$ of each cluster $i\in [1,k]$. The average cluster size $\bar{s}$ is around 300. The \#128-th cluster is a huge miscellaneous cluster and contains 2,948 images (i.e., about 6\% of all the images). This cluster seems to contain all the ``outsiders'' around the edge of the logo image distribution. Therefore, this miscellaneous cluster will not show any clear trend within it and is excluded from the later analysis.\par

\begin{figure}[!p]
\begin{center}
\begin{minipage}[b]{0.32\textwidth}
    \centering
    \includegraphics[width=\linewidth]{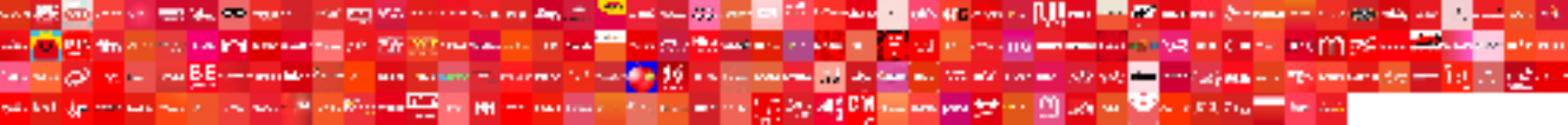}\\
    \includegraphics[width=\linewidth]{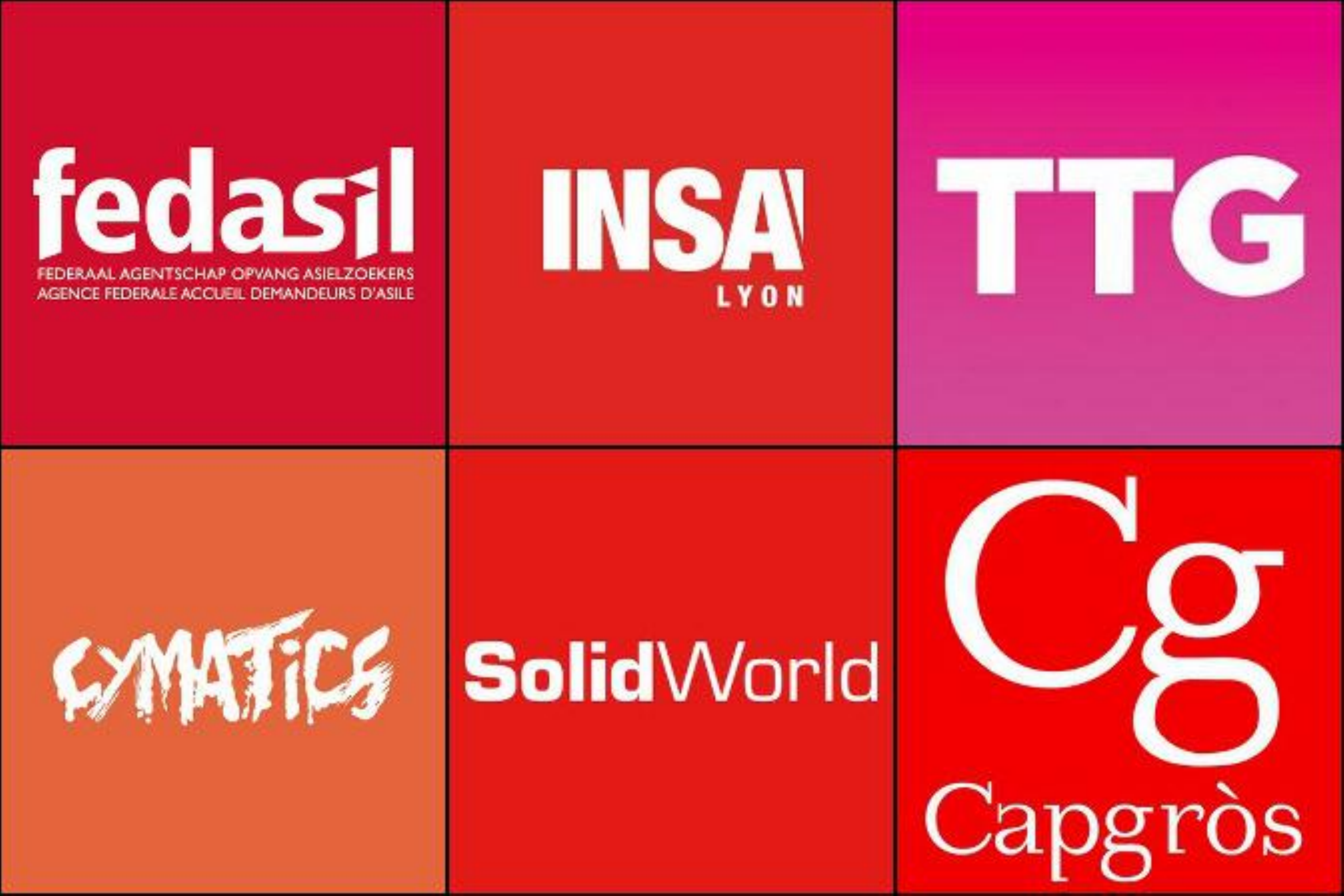}\\[-1mm]
    Cluster \#1
\end{minipage}
\begin{minipage}[b]{0.32\textwidth}
    \centering
    \includegraphics[width=\linewidth]{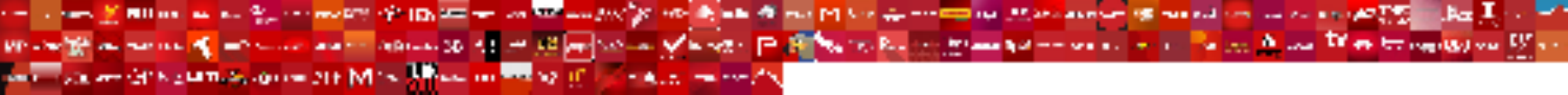}\\
    \includegraphics[width=\linewidth]{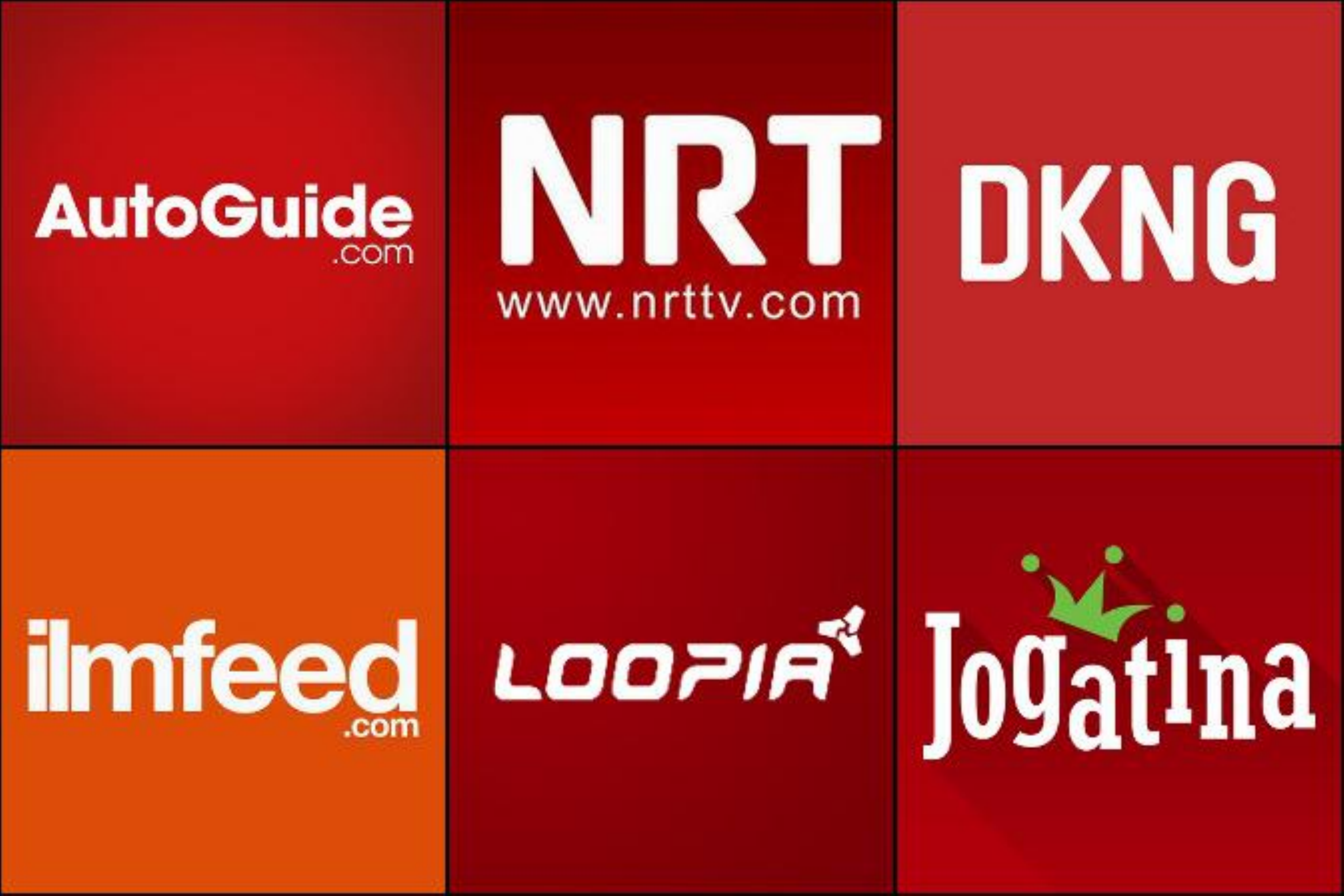}\\[-1mm]
    Cluster \#2
\end{minipage}
\begin{minipage}[b]{0.32\textwidth}
    \centering
    \includegraphics[width=\linewidth]{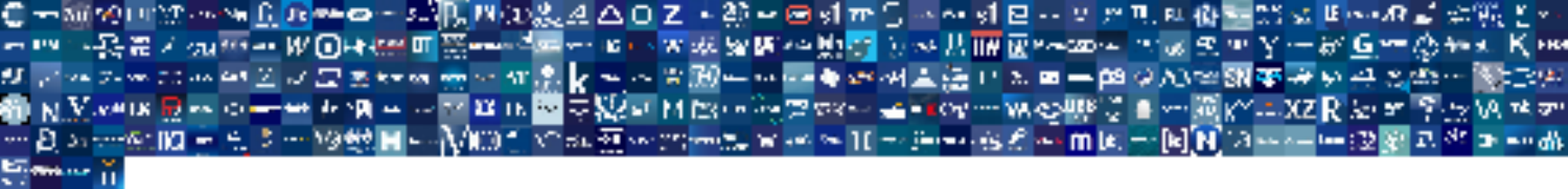}\\
    \includegraphics[width=\linewidth]{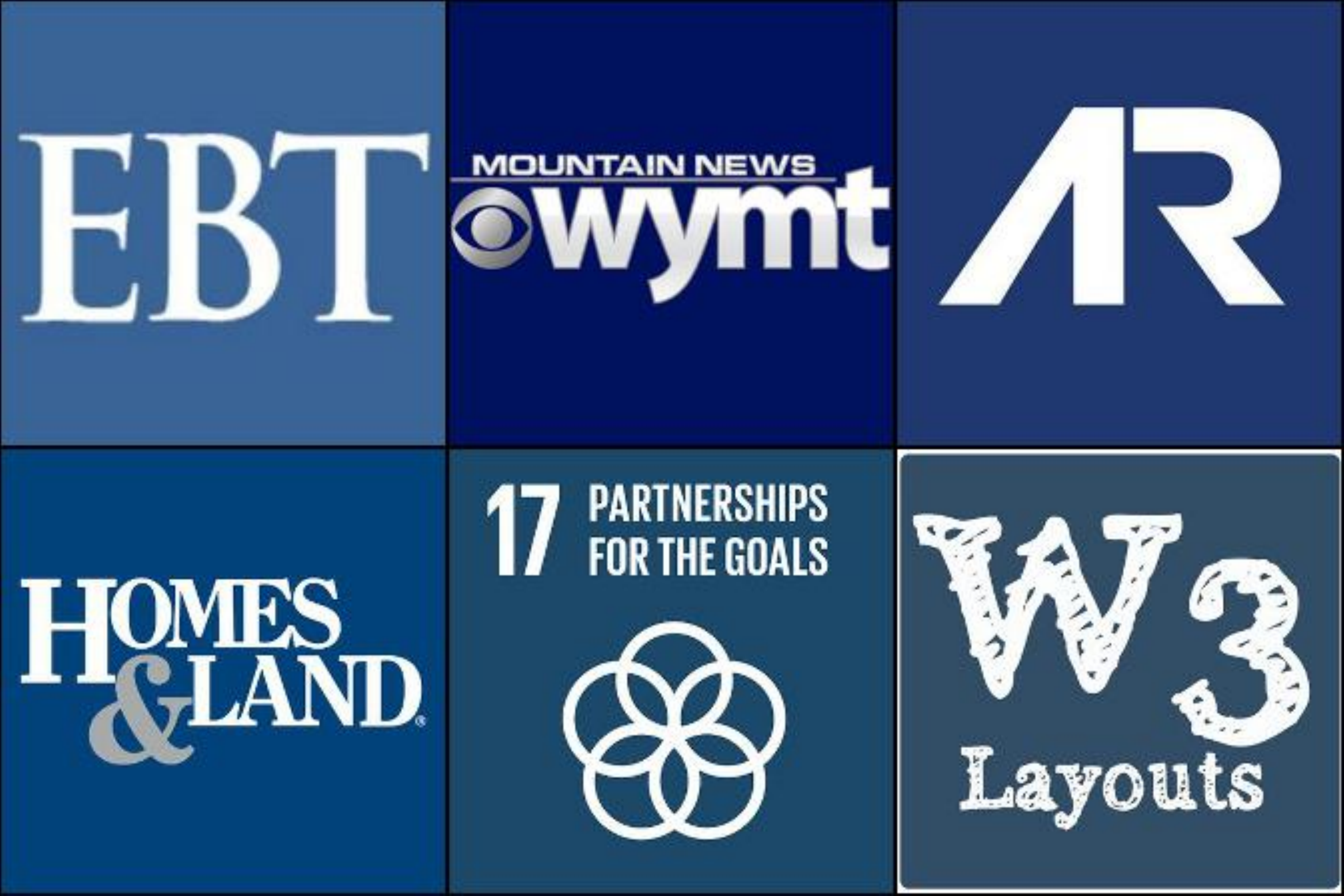}\\[-1mm]
    Cluster \#3
\end{minipage}\medskip \\
\begin{minipage}[b]{0.32\textwidth}
    \centering
    \includegraphics[width=\linewidth]{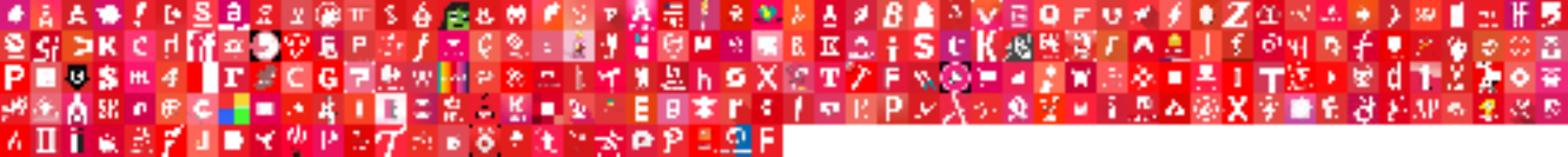}\\
    \includegraphics[width=\linewidth]{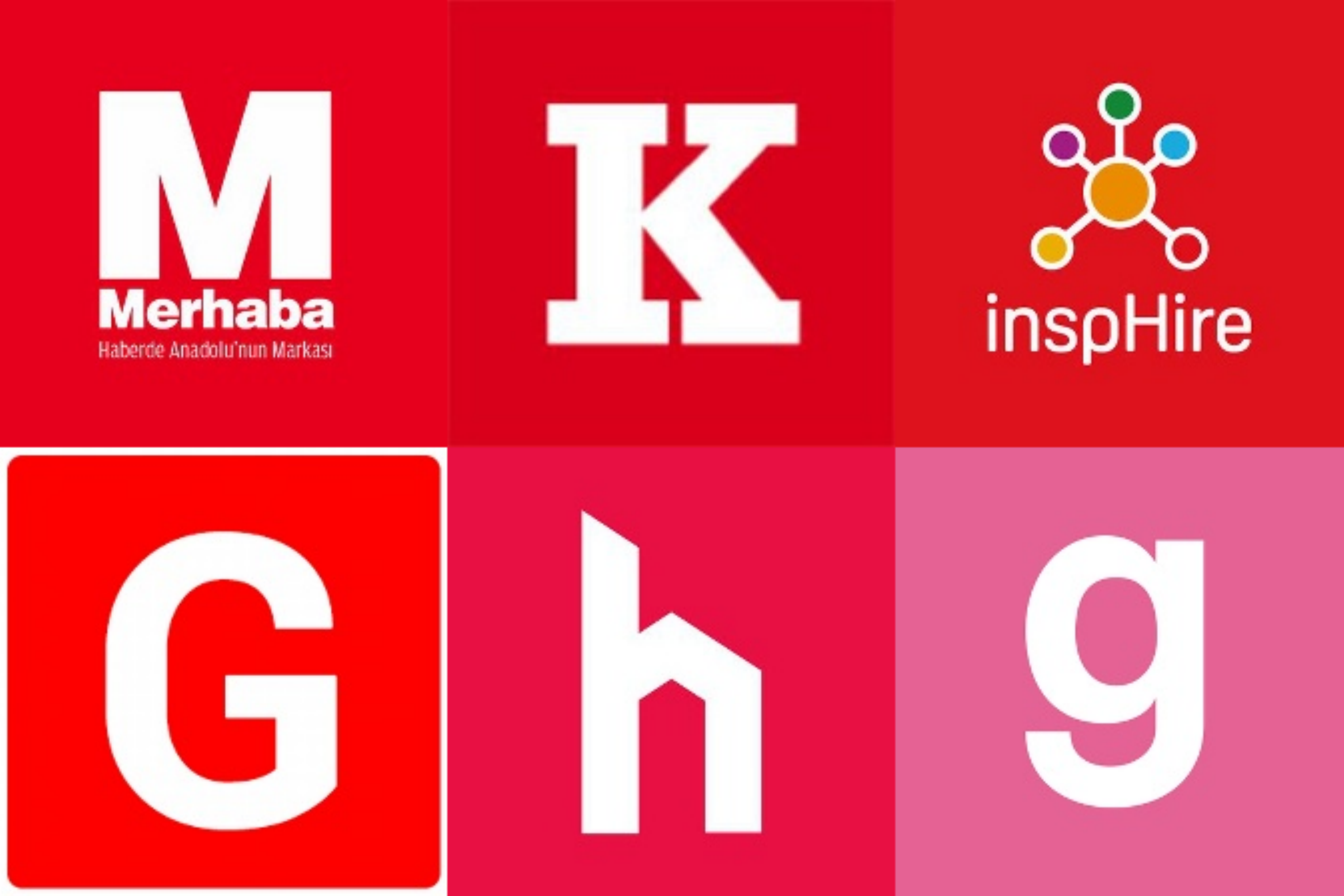}\\[-1mm]
    Cluster \#4
\end{minipage}
\begin{minipage}[b]{0.32\textwidth}
    \centering
    \includegraphics[width=\linewidth]{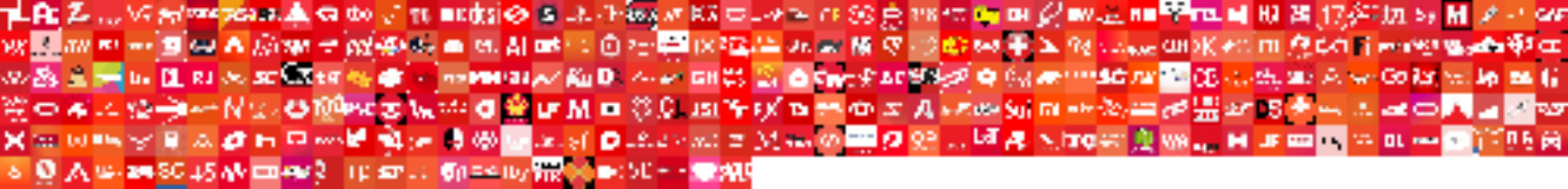}\\
    \includegraphics[width=\linewidth]{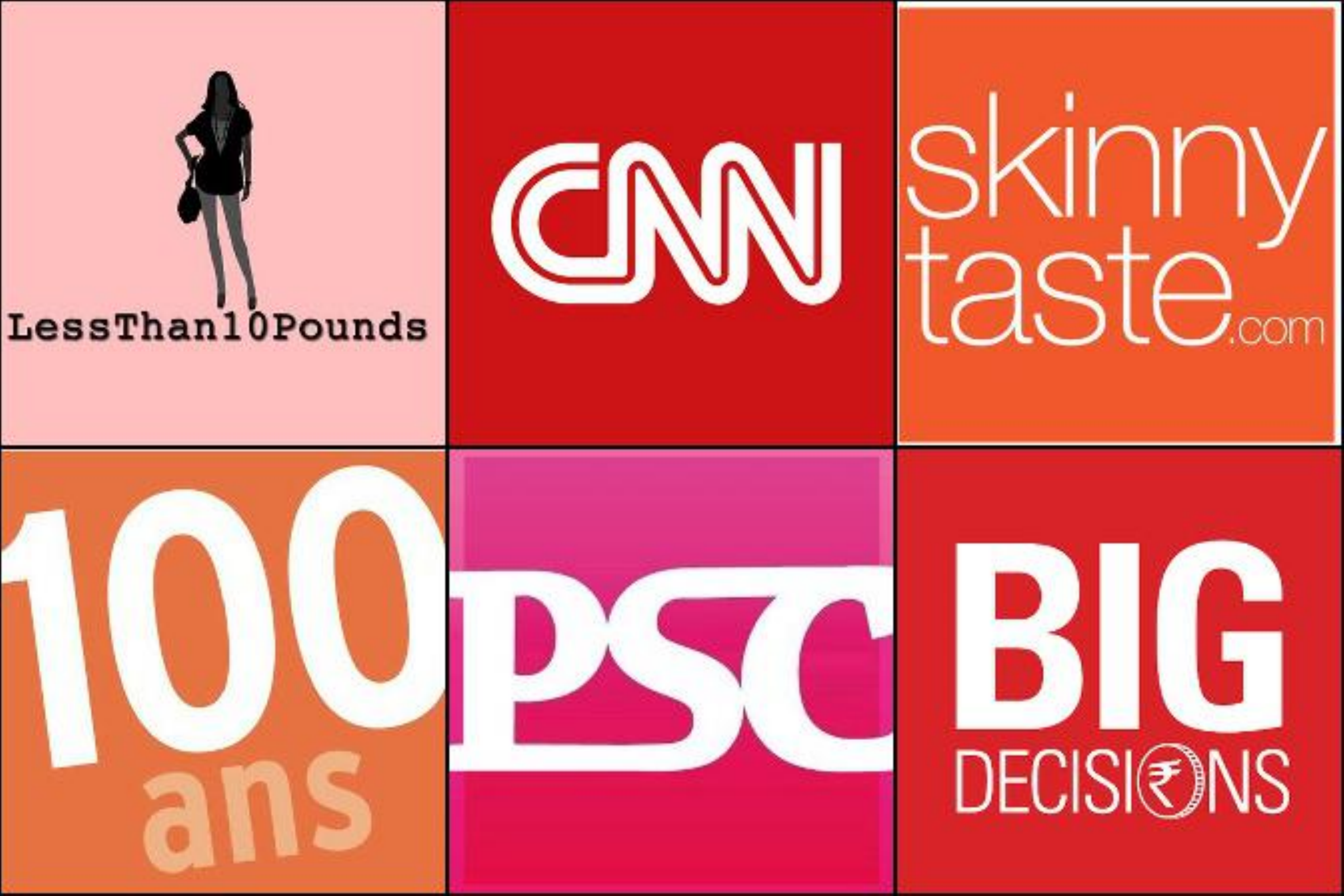}\\[-1mm]
    Cluster \#5
\end{minipage}
\begin{minipage}[b]{0.32\textwidth}
    \centering
    \includegraphics[width=\linewidth]{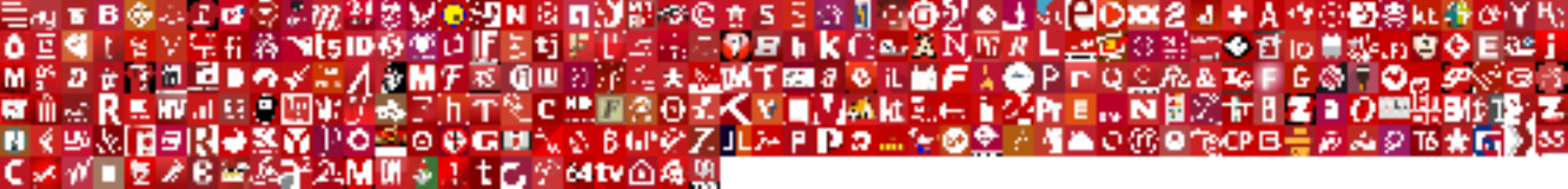}\\
    \includegraphics[width=\linewidth]{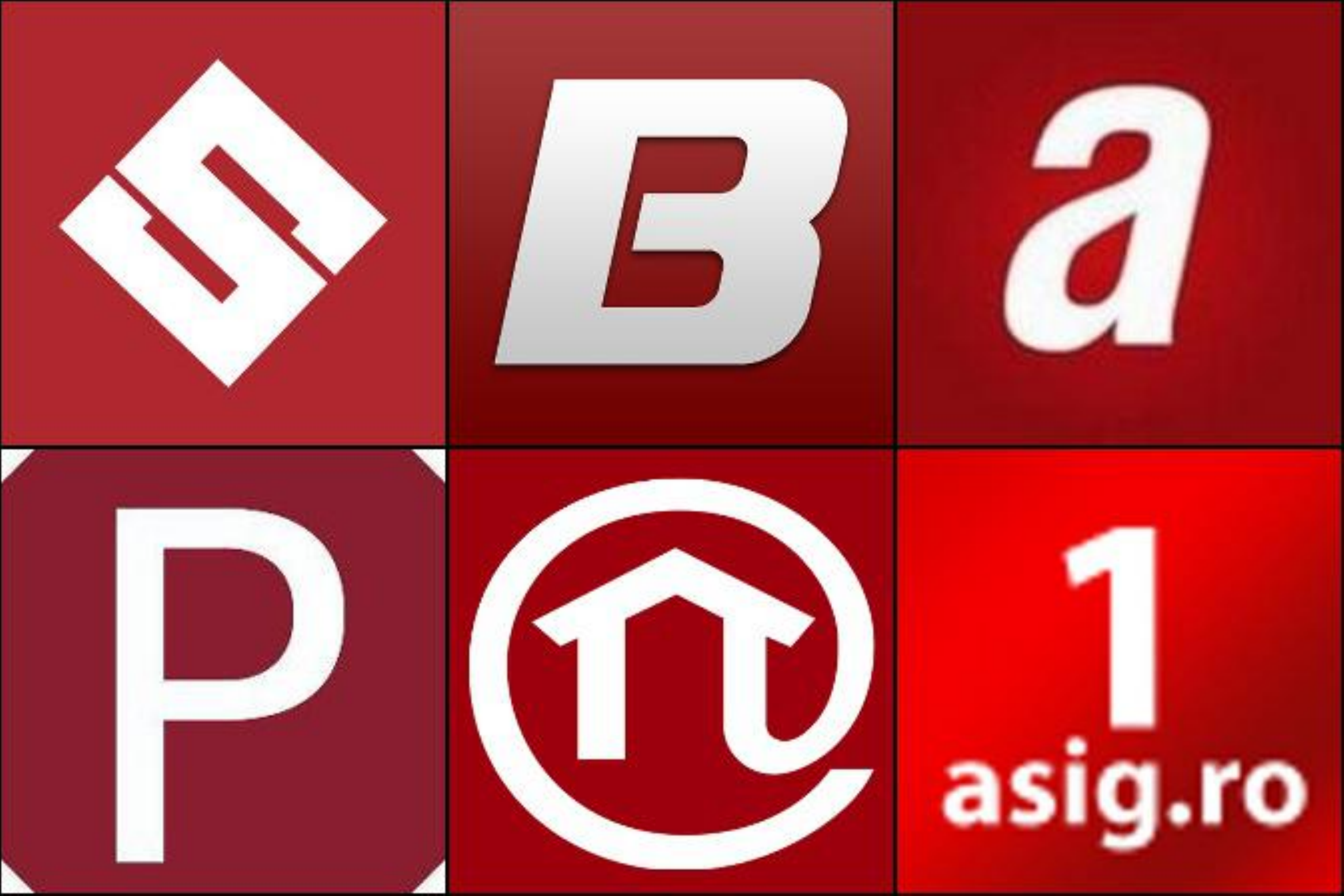}\\[-1mm]
    Cluster \#6
\end{minipage}\medskip \\
\begin{minipage}[b]{0.32\textwidth}
    \centering
    \includegraphics[width=\linewidth]{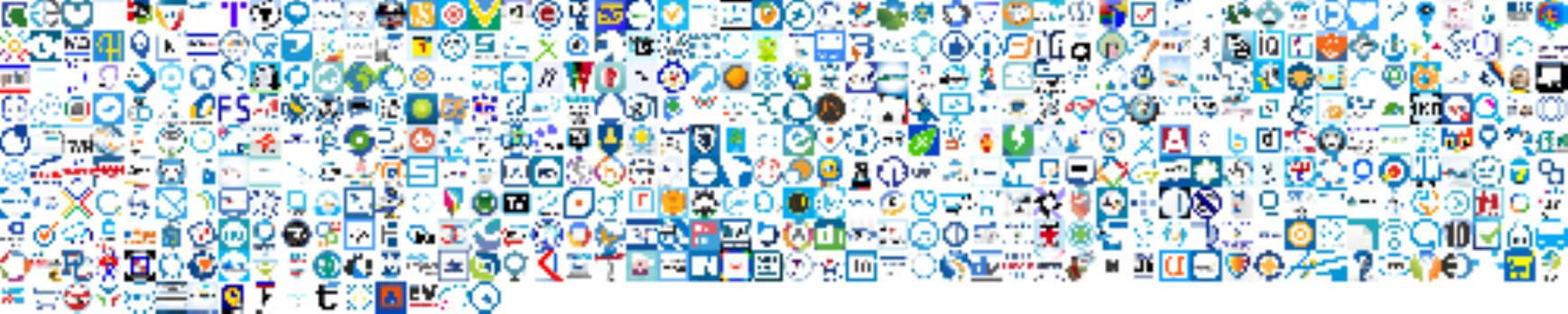}\\
    \includegraphics[width=\linewidth]{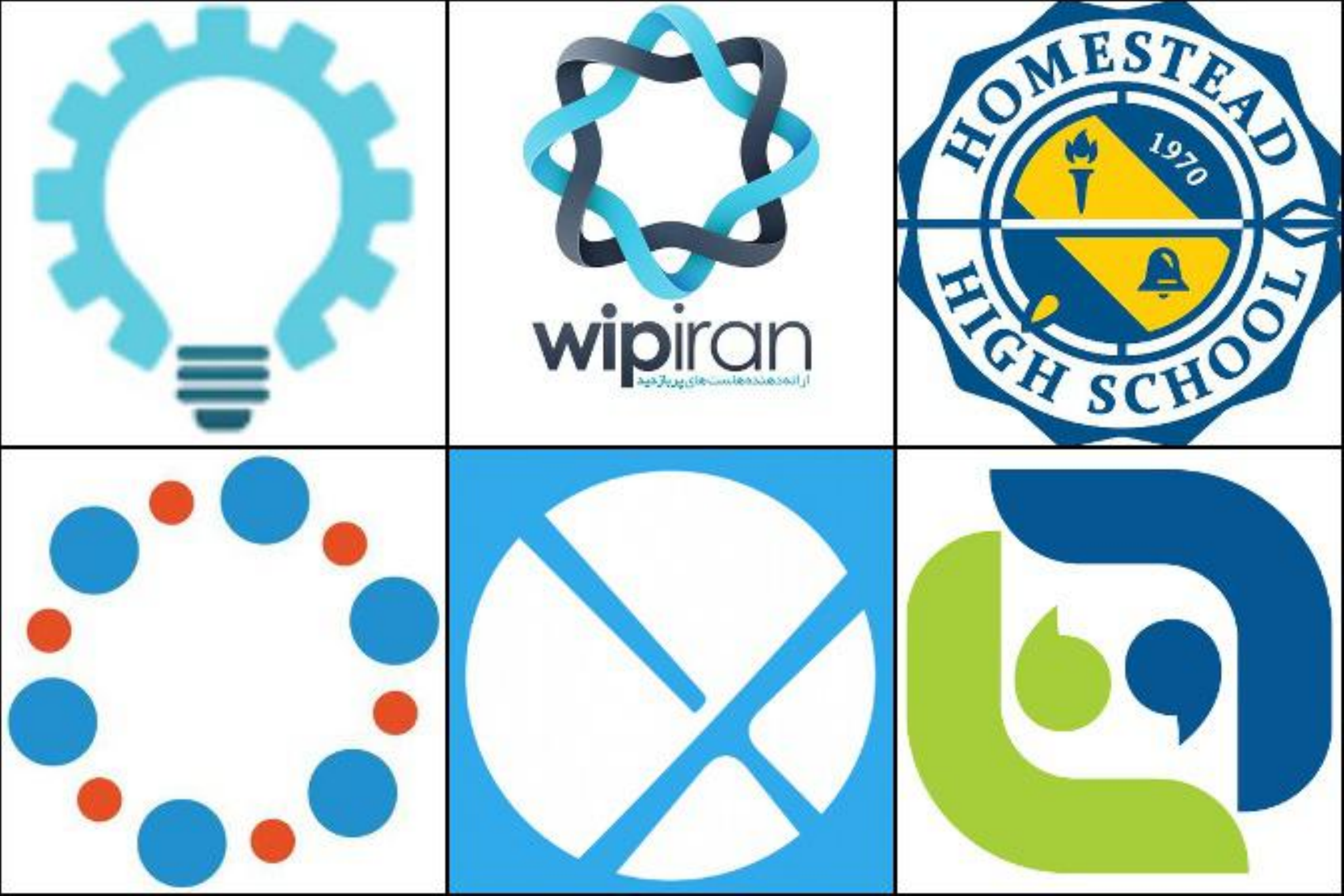}\\[-1mm]
    Cluster \#125
\end{minipage}
\begin{minipage}[b]{0.32\textwidth}
    \centering
    \includegraphics[width=\linewidth]{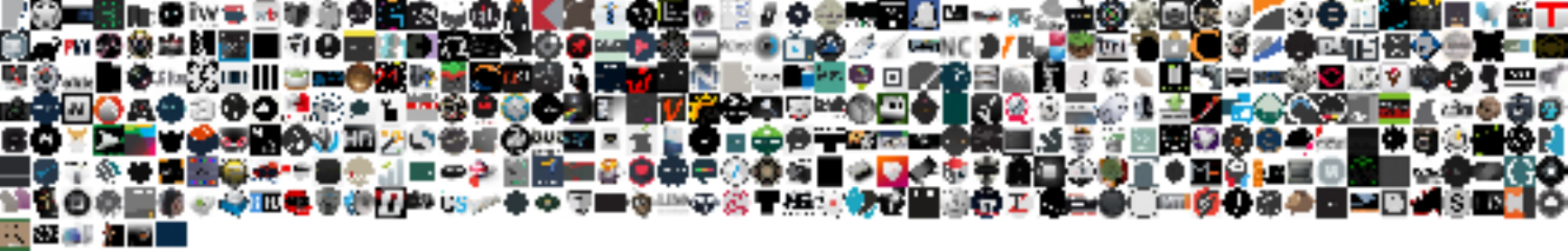}\\
    \includegraphics[width=\linewidth]{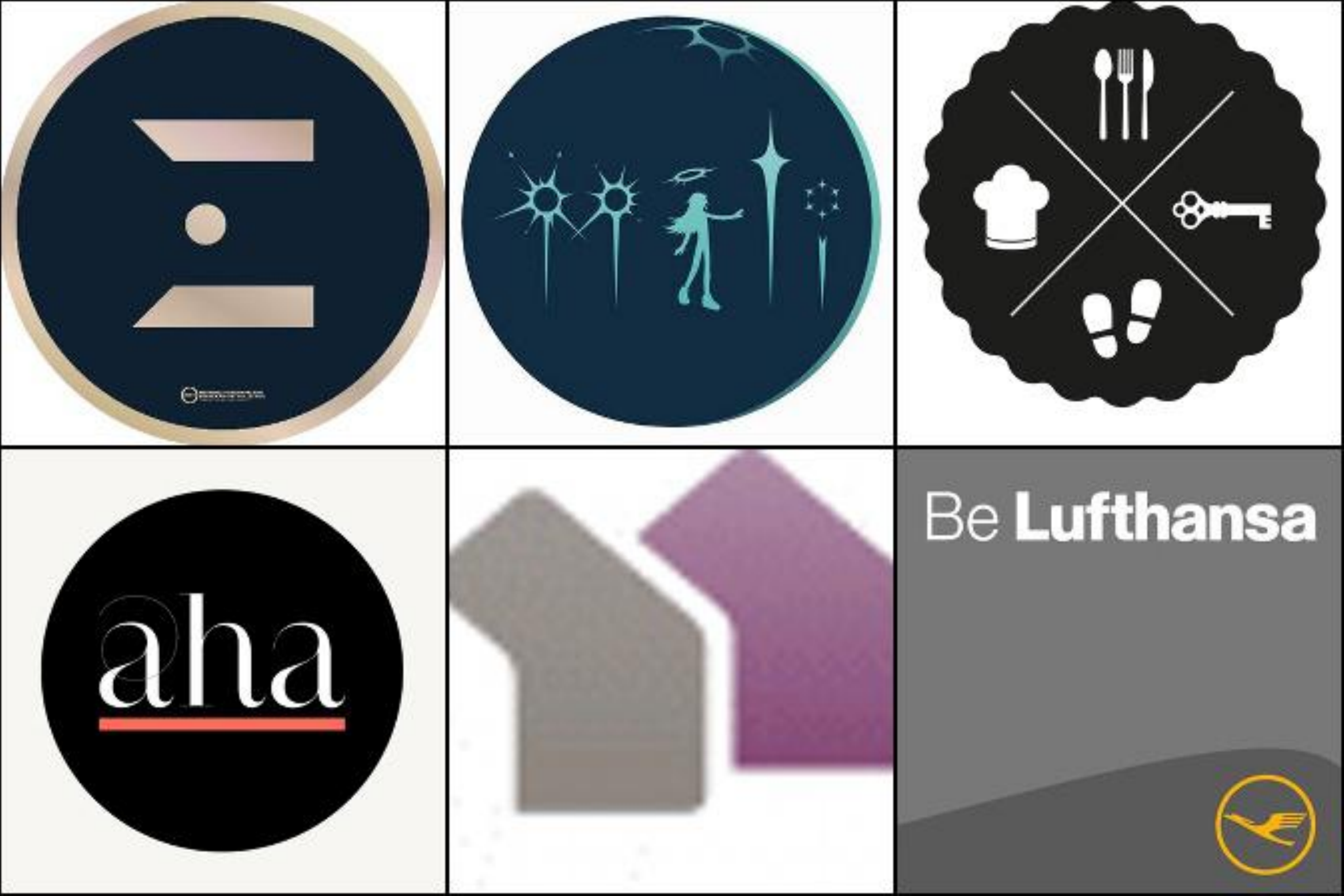}\\[-1mm]
    Cluster \#126
\end{minipage}
\begin{minipage}[b]{0.32\textwidth}
    \centering
    \includegraphics[width=\linewidth]{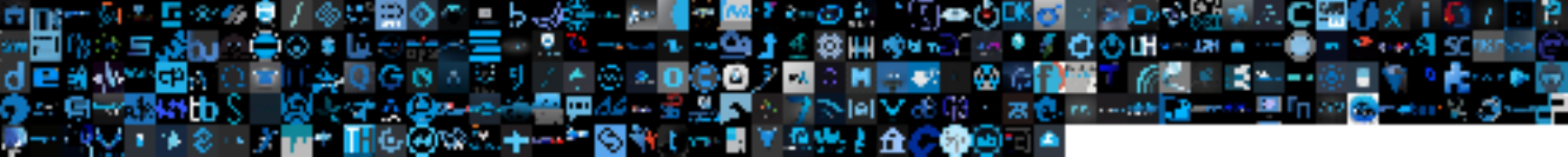}\\
    \includegraphics[width=\linewidth]{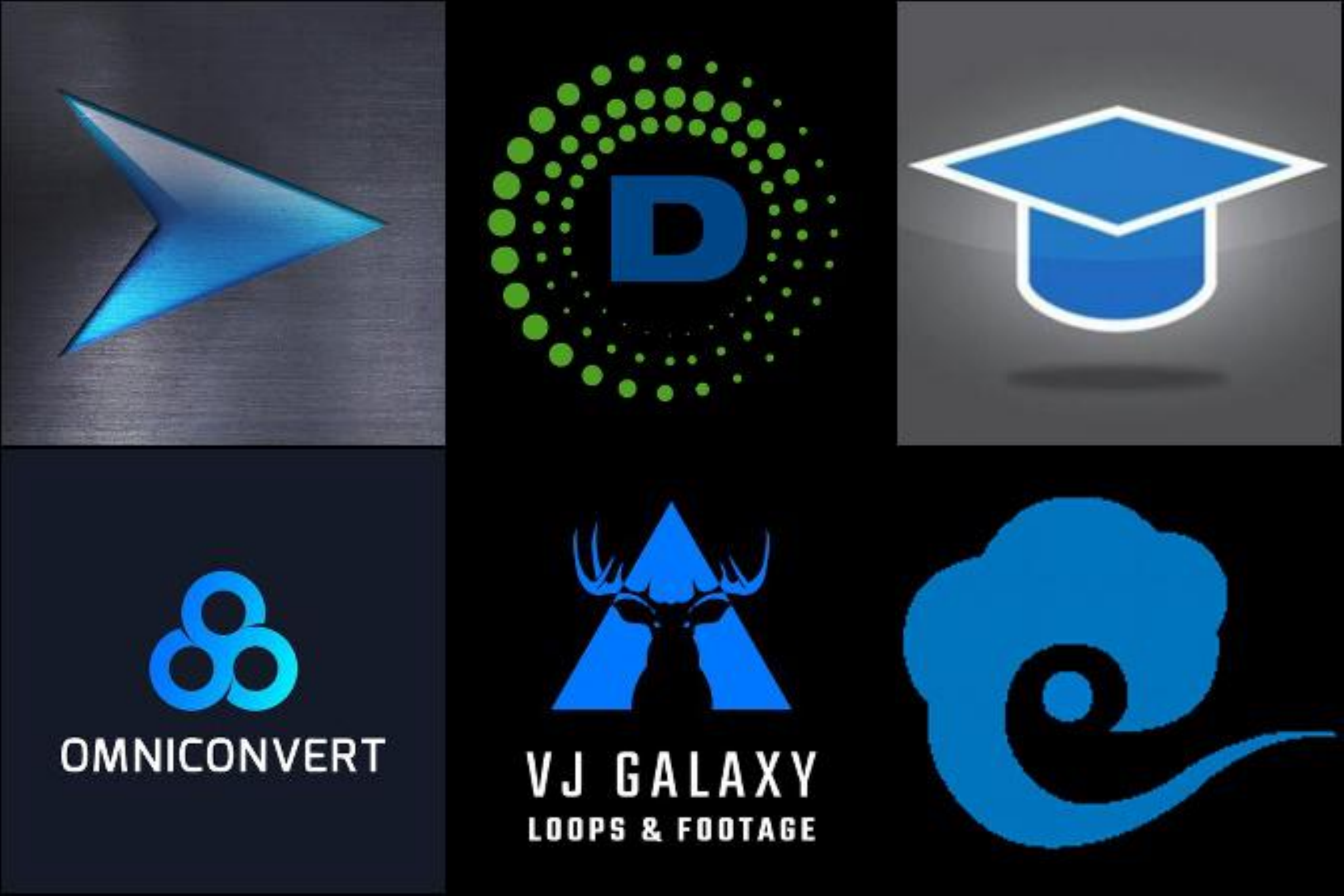}\\[-1mm]
    Cluster \#127
\end{minipage}\medskip \\
\begin{minipage}[b]{0.32\textwidth}
    \centering
    \includegraphics[width=\linewidth]{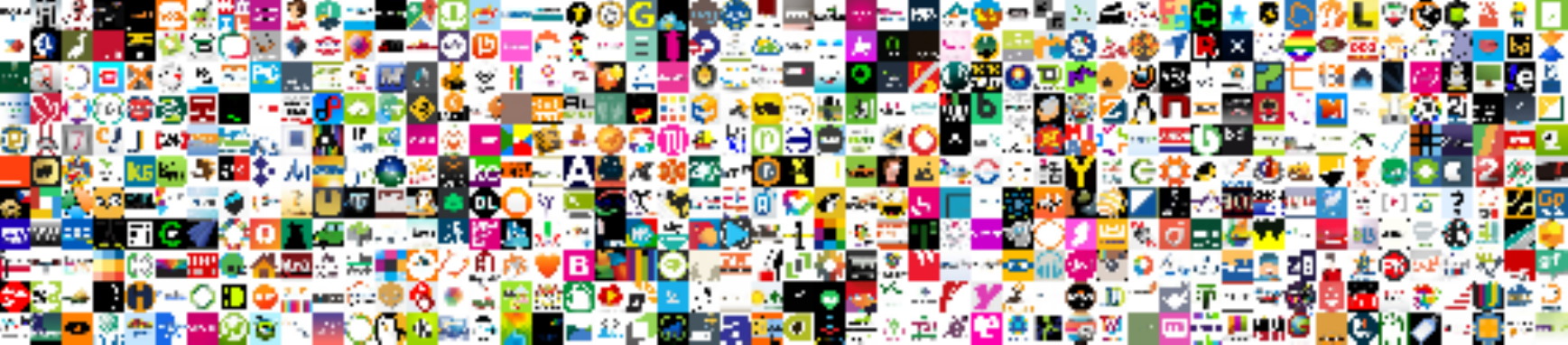}\\
    \includegraphics[width=\linewidth]{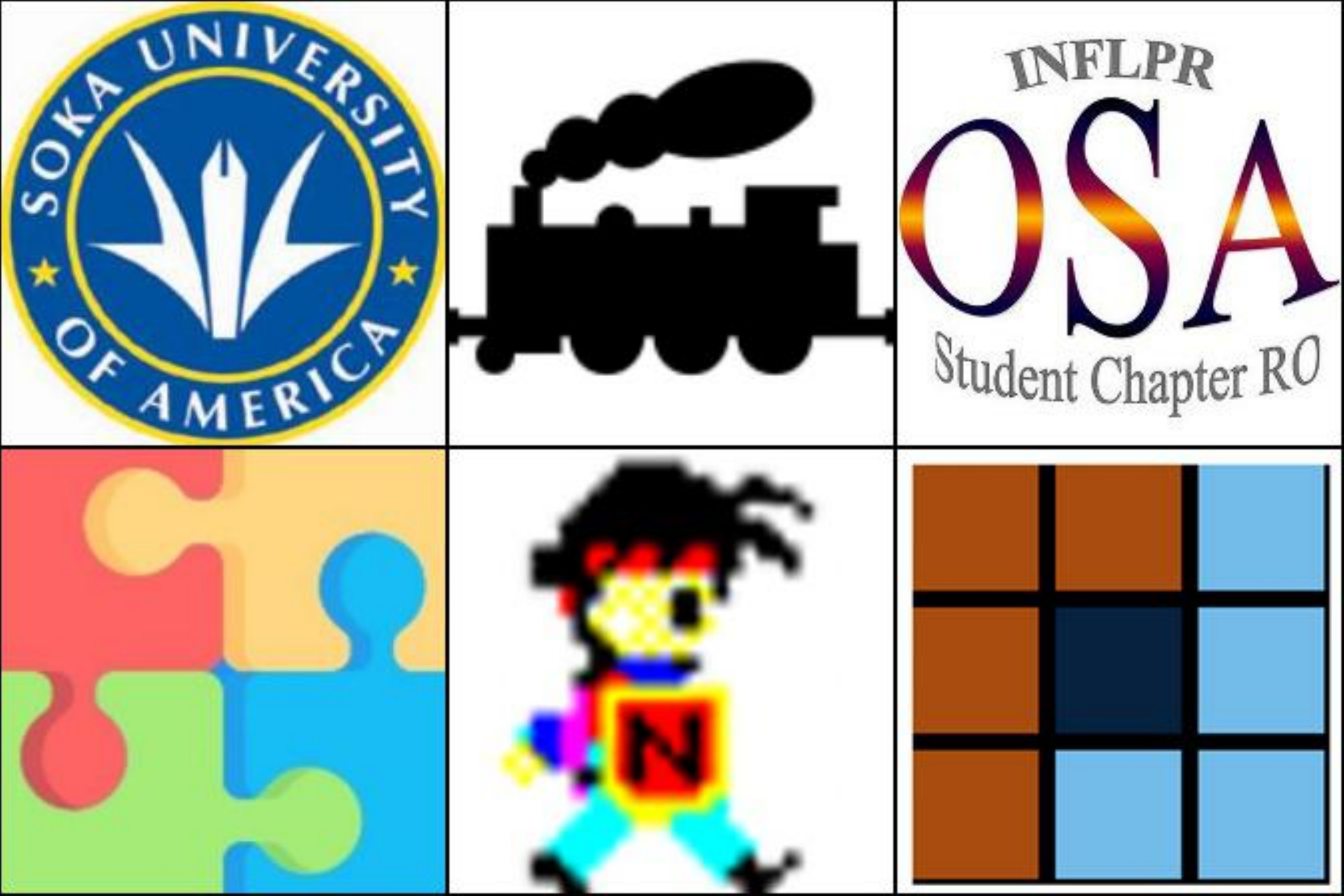}\\[-1mm]
    Cluster \#128
\end{minipage}
\begin{minipage}[b]{0.32\textwidth}
    \centering
    \includegraphics[width=\linewidth]{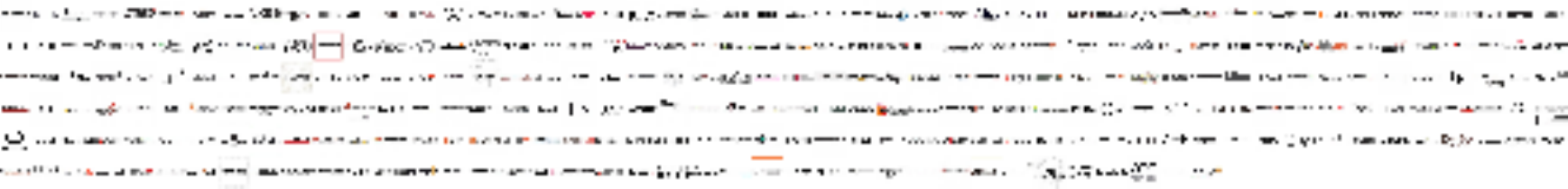}\\
    \includegraphics[width=\linewidth]{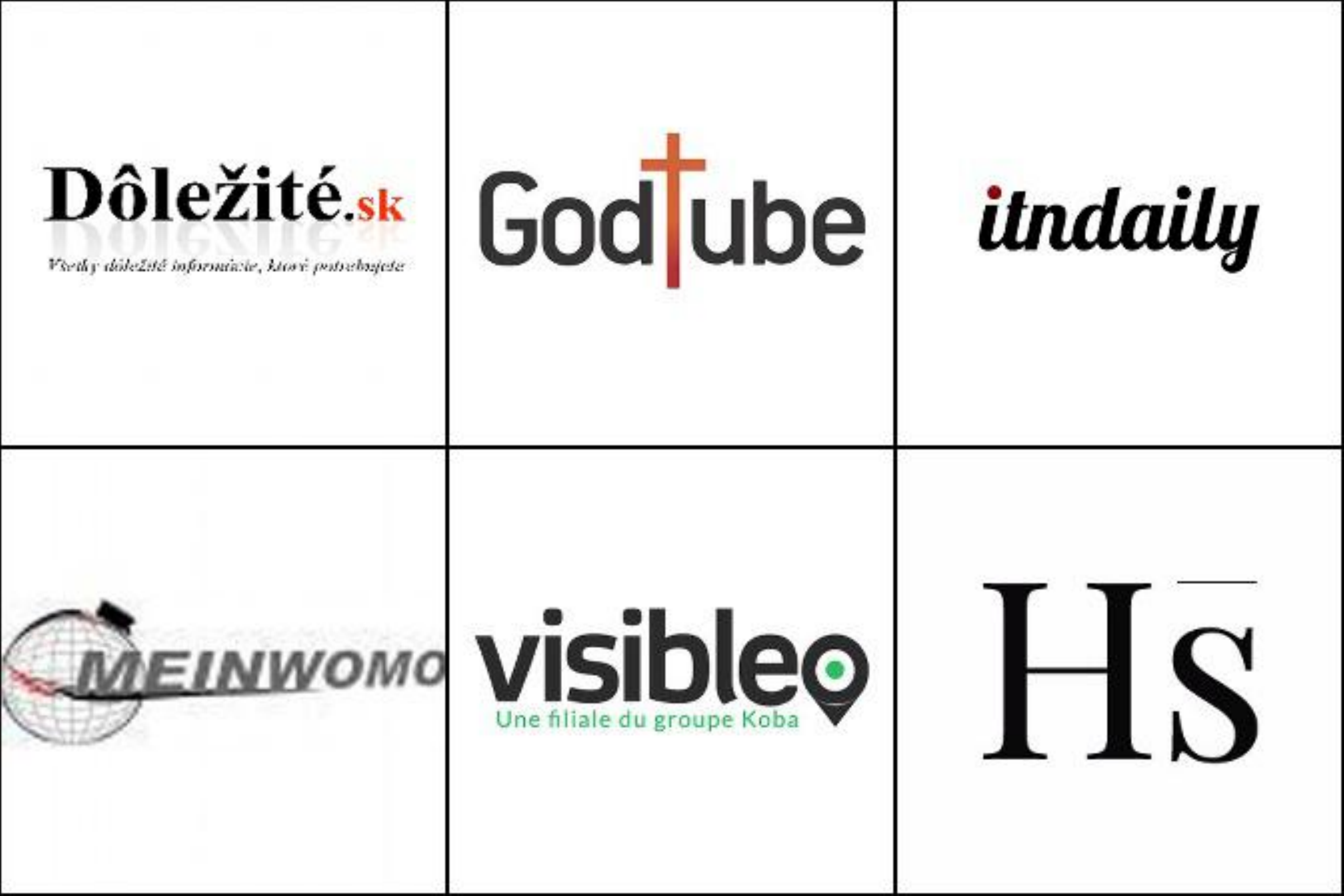}\\[-1mm]
    Cluster \#96
\end{minipage}
\begin{minipage}[b]{0.32\textwidth}
    \centering
    \includegraphics[width=\linewidth]{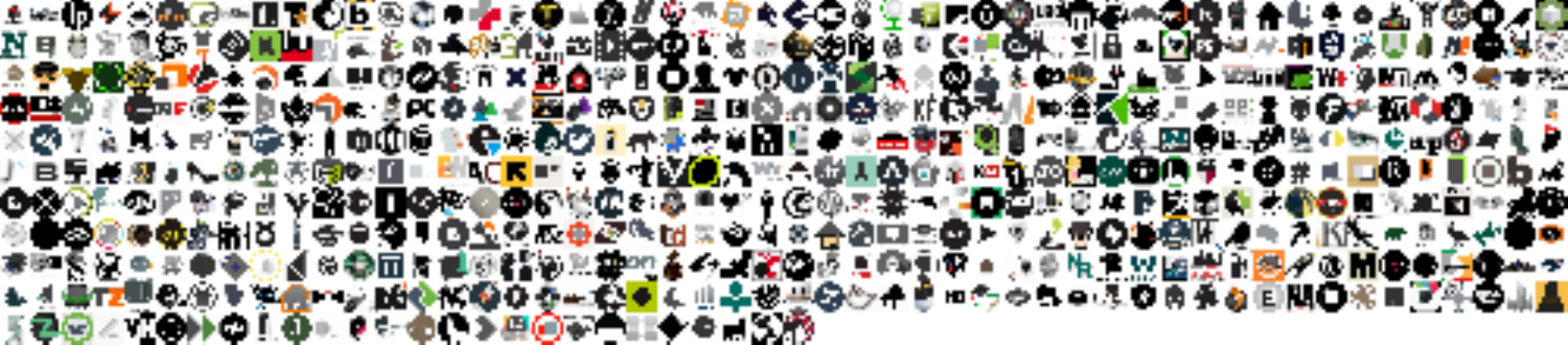}\\
    \includegraphics[width=\linewidth]{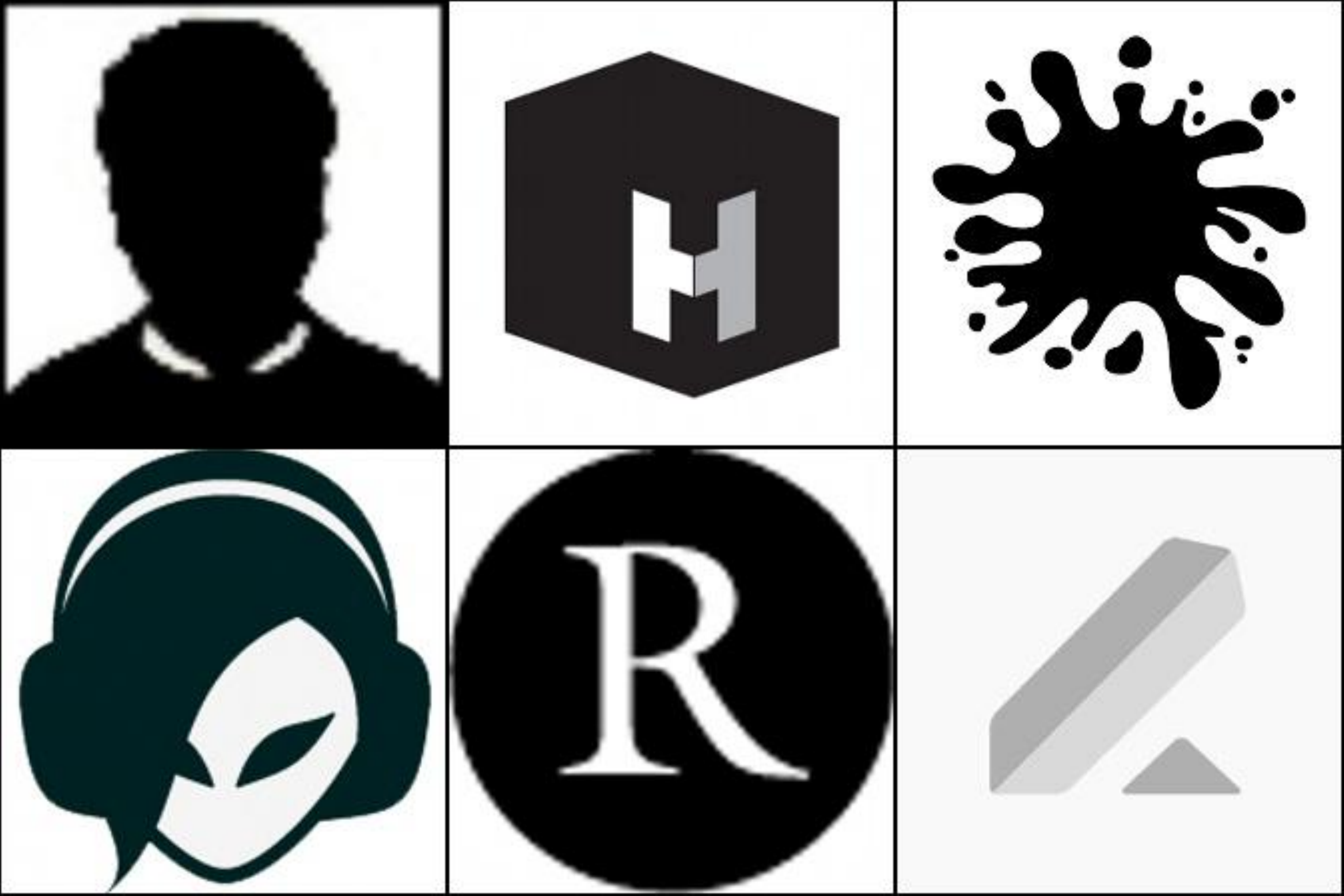}\\[-1mm]
    Cluster \#113
\end{minipage}\\
    \caption{Logo examples of several representative clusters. For each cluster, the tiny thumbnail images of all the logo images in the cluster is shown above the examples, for grasping the general color trend of the cluster. Those thumbnail images are also shown in Fig.~\ref{fig:cluster-wise-area-distribution.}(d). Only in the large miscellaneous cluster \#128, not all images are shown.}
    \label{fig:cluter-samples}
\end{center}
\end{figure}

%------------------------------------------------------
\subsection{The number of followers at each cluster}
Fig.~\ref{fig:cluster-wise-area-distribution.}~(b) shows the box plot showing the number of followers of the logos that belong to each cluster. Let $x_{i,j}$ denote the $j$th logo image of the $i$th cluster ($j\in [1,s_i]$) and $f_{i,j}$ is the number of its followers.
In this figure (also (a) and (c)), the clusters are sorted in the descending order of the median of the number of followers; this means that the condition $f_i\geq f_{i-1}$ always holds for all $i$, where $f_i = \mathrm{Med}_jf_{i,j}$. Note that the outlier dots are not plotted in the box plot for better visibility. \par
Since many logo images have around $f_{i,j} = 10^2\sim 10^4$ followers
as already shown by Fig.~\ref{fig:popu_dist}, the medians $\{f_i\}$ (indicated by orange) are also around $10^3$; however, they are still different. For example, the median $f_1\sim 3,000$, whereas $f_{127}\sim 500$. 

%-----------------------------------------------------
\subsection{The text area ratio of each cluster}
Fig.~\ref{fig:cluster-wise-area-distribution.}~(c) is the box plot showing the text area ratios of the logos that belong to each cluster. Let $t_{i,j}$ denote the text area ratio of the $j$th logo in the $i$th cluster and $t_i = \mathrm{Med}_j t_{i,j}$. The median $t_i$ is depicted as a red dot in (c). \par
Some clusters have larger $t_i$ (i.e., more logotypes), whereas some have smaller $t_i$ (i.e., more logo symbols). This means that DeepCluster forms clusters while considering the text areas in logo images. At the same time, there is no cluster whose box height is almost zero; this means that no logotype-only cluster and no logo symbol-only cluster.\par
Note again that the clusters are arranged in the descending order of $f_i$ and not $t_i$ in Fig.~\ref{fig:cluster-wise-area-distribution.}. Therefore, no monotonicity condition holds for $t_i$. However, in Section~\ref{sec:correlation}, we will see that there is a weak trend that 
$t_i$ tends to be larger than $t_{i'}$ for $i<i'$. This suggests the followers $f_i$ and texts $t_i$ are weakly correlating.

%------------------------------------------------------
\subsection{Logos in several clusters\label{sec:cluster-members}}
Fig.~\ref{fig:cluter-samples} shows the logo image examples from several clusters. For each cluster, six representative examples are shown. A tiny thumbnail showing all the logos in the cluster is also shown above the examples, for capturing the trend (especially  color) of the cluster.
\par
Clusters \#1, \#2, $\ldots$, \#6 are the six clusters with the highest $f_i$, and Clusters \#125, \#126 and \#127 are the lowest. From those examples, we can observe that the DeepCluster composes its clusters by extracting various features from logo images. \par
Especially, the top six clusters have very consistent color and shape styles within the cluster. The thumbnail images of those clusters prove this color homogeneity within each cluster.
Except for \#4, logos with a red background and a white large text are common among them. Although we do not give any prior knowledge about text areas to DeepCluster, it could find ``text-like'' elements as features for the clustering.\par
It is interesting to observe that red logos indeed occupy most of the top clusters. Since the clusters' order is determined by the median $f_i$, we cannot say the red logos {\it always} correlate the companies with high popularity. However, at least the cluster-wise average (by means of median) shows this trend.  Fig.~\ref{fig:cluster-wise-area-distribution.}(d) supports this trend.
In contrast, the logo images from Clusters \#125-127 (having the lowest $f_i$) often show finer structures and pictorial elements (rather than texts) in their design.\par
Cluster \#128 is the miscellaneous cluster with an extremely large number of logos (2,948). As expected from its size, it contains logos with various appearances. Since DeepCluster is a classifier-based clustering method, it may realize such a ``rejection'' class. This can be confirmed from its colorful thumbnail image.
We excluded this cluster from the trend analysis because it seems to have no clear trend.\par
Clusters \#96 and \#113 are the clusters with the maximum and minimum $f_i$, respectively. The former contains logotypes; the latter contains logo symbols and mixed logos where texts are often surrounded by a large background element.

%-----------------------------------------------------
\subsection{Cluster-wise correlation analysis between the number of followers and the text area ratio\label{sec:correlation}}
The red bar in Fig.~\ref{fig:cluster-wise-area-distribution.}~(c) shows the median of text area ratios of logos in each cluster. The red dots seem to fluctuate from Cluster \#1 (leftmost) to Cluster \#127 (rightmost); however, careful observation will find the trend that the text area ratio decreases along with the cluster ID. The blue curve is a smoothed transition of the median values (red dots) of text area ratios.  For the smoothing, the median filter with a width of 13 is applied in the Cluster-ID direction, to catch the trend of the curve robustly.\par
\par
Considering that cluster ID is relative to the number of followers, the slope of the blue line catches the weak trend that {\em the number of followers correlates positively with the text area ratio}. In other words, the more followers a company has, the more texts its logo contains. Of course, this correlation does not always hold --- however, this cluster-wise median-based robust estimation reveals this weak trend. It should be emphasized that this conclusion coincides with our previous analysis in Section~\ref{sec:two-hists} with Fig.~\ref{fig:follower-compare-1}.

\begin{figure}[t]
    \centering
    \begin{minipage}{0.45\textwidth}
        \centering
    \includegraphics[width=\linewidth]{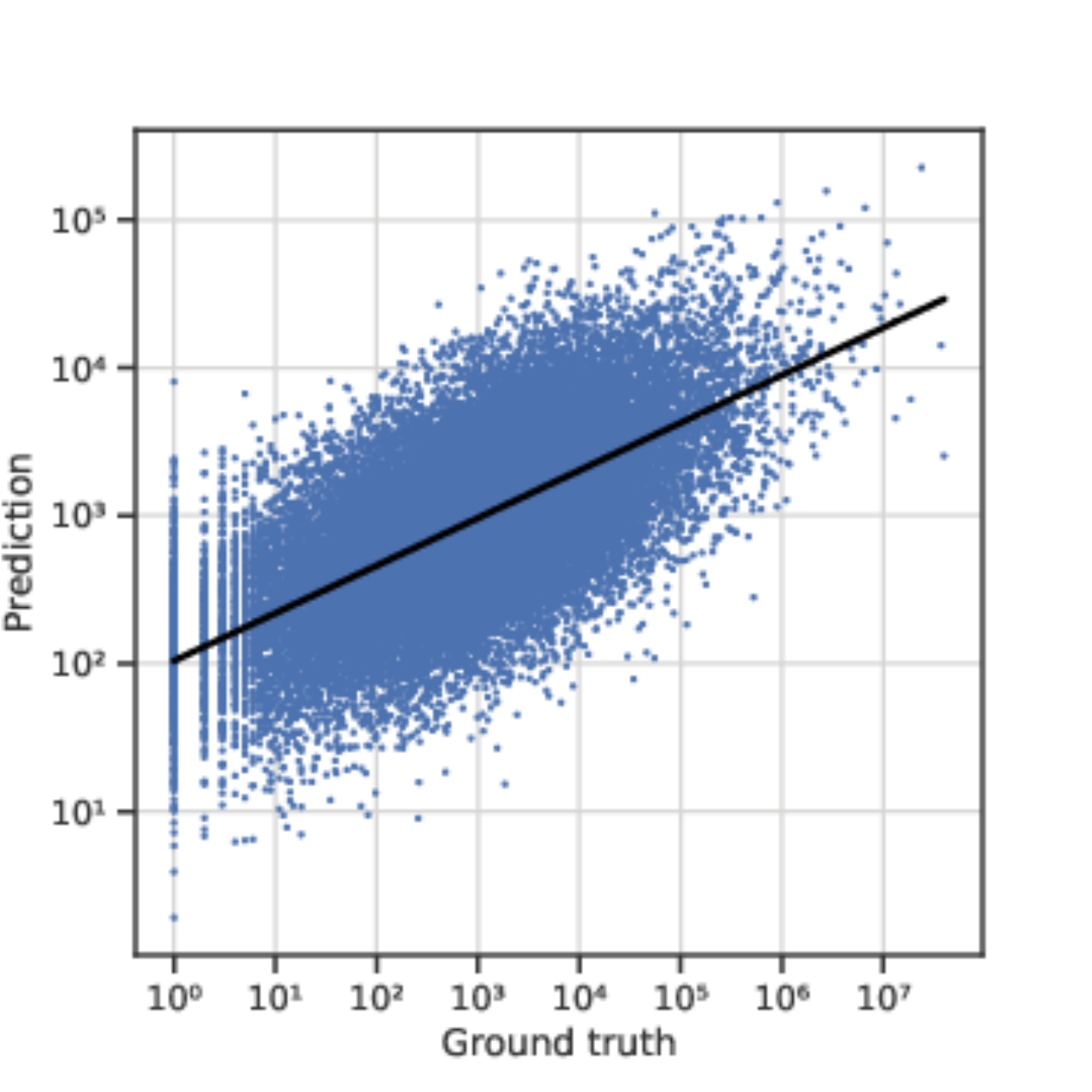}\\[-1mm]
    (a) Training samples.
    \end{minipage}
    \begin{minipage}{0.45\textwidth}
        \centering
    \includegraphics[width=\linewidth]{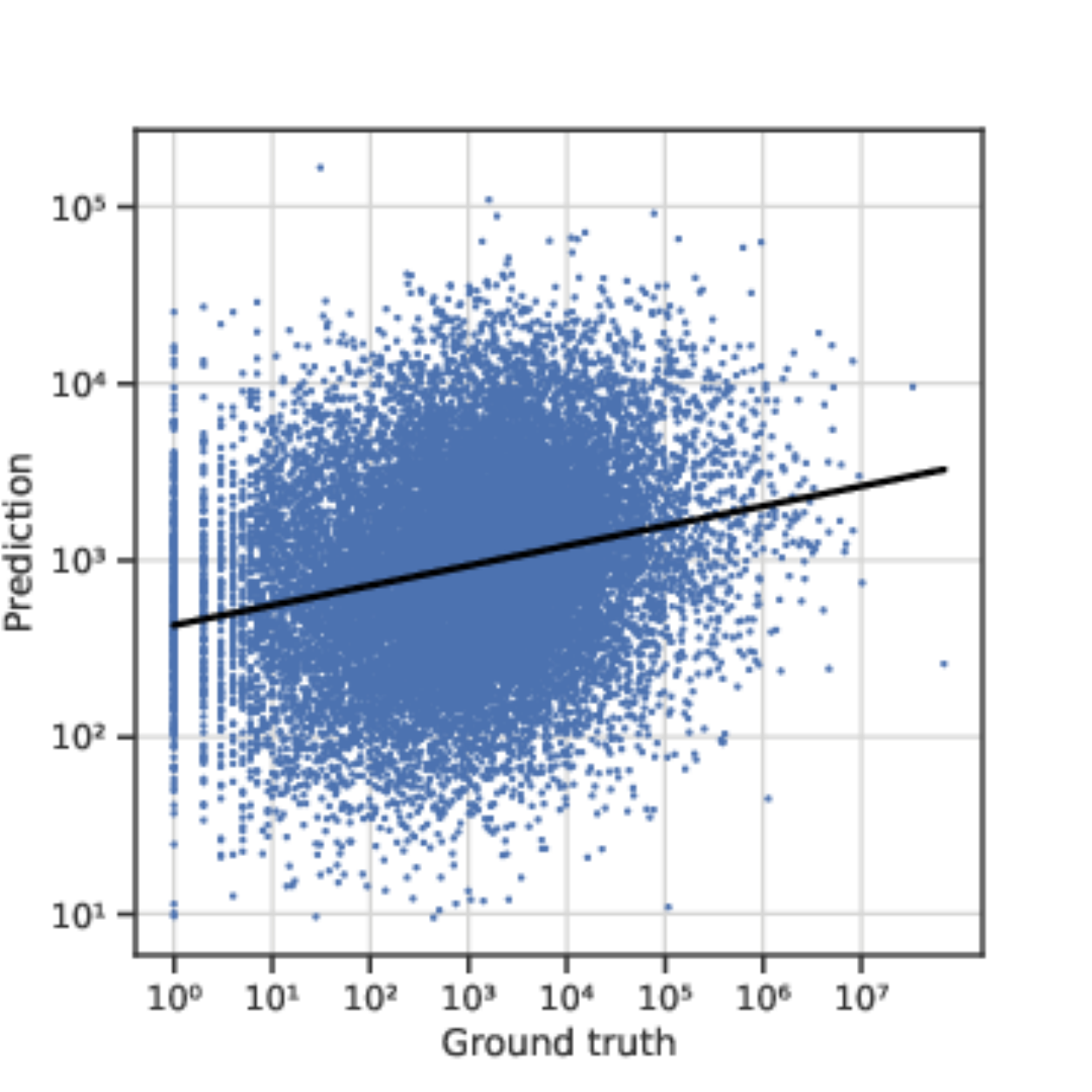}\\[-1mm]
      (b) Test samples.
    \end{minipage}  
    \caption{Prediction result of the number of followers from logo images.}
    \label{fig:regression result}
    \vskip -4mm
\end{figure}
%%%%%%%%%%%%%%%%%%%%%%%%%%%%%%%%%%%%%%%%%%%%%%%%%%%%%%%%%%%
\section{Analysis 3: Estimation of the Number of Followers from Logo Image}

\subsection{Regression-based estimation}
In this section, we conduct a challenging task of estimating the number of followers from a logo image by regression analysis. This task is challenging because the number of followers depends on so many factors, including the company's product, the market trends, the size of the company, etc. However, we tackle this task because of two positive evidences. First, the recent deep regression models show their powerful ability to deal with extremely nonlinear regression tasks. Second, as we observed already, we could catch the weak correlation between the text area ratio and the number of followers, by the cluster-wise analysis in Section~\ref{sec:correlation}. \par
The deep regression is used to estimate the number of followers for the input log image. All logo images are re-scaled to $224\times 224$ pixels. 
The network structure is DenseNet-169, and the standard MSE loss function is used.
For training, validating, and testing, 30,000, 10,000, and 20,000 logo images randomly selected from LLD-logo are used, respectively. The standard early stopping rule (no improvement on the validation accuracy for 10 epochs) is used to terminate the training process. During training and testing, the number of followers is treated in a logarithmic manner due to the same reason of Section~\ref{sec:data}.
\par
Figs.~\ref{fig:regression result} (a) and (b) show the scattered plots of the ground-truth (GT) and prediction (PR) for the training samples and test samples, respectively. The black line is the linear regression result between GT and PR. Surprisingly, the linear regression result shows that 
GT and PR are clearly correlating; this means that the prediction of the number of followers is possible to some extend. (At least, it 
is not impossible.) The Pearson correlation coefficients between GT and PR for the training and test sets are about 0.64 and 0.23, respectively. The test correlation is not very strong but it is enough to confirm the existence of the correlation.
The $p$-values are almost zero and satisfy $p$ for both sets, and thus we can also confirm that GT and PR are actually correlated. \par

\subsection{Ranking-based estimation}
As another formulation of the estimation task, we conduct a learning-to-rank experiment with the idea of RankNet~\cite{Burges2005}. The idea of RankNet is 
to determine a ranking function $r(x)$ that satisfies the condition $r(x_1)>r(x_2)$ when $x_1$ should be ranked higher than $x_2$. In our case, $x$ represents a logo image and if the logo image $x_1$ more followers than $x_2$, the condition should be satisfied. If we have an accurate ranking function $r(x)$, it can predict a {\it relative} popularity, such as $r(x_1)>r(x_2)$ or $r(x_1)<r(x_2)$, for a given pair of images $\{x_1, x_2\}$.
\par
RankNet realizes a nonlinear ranking function by its neural network framework. The original RankNet~\cite{Burges2005} is implemented by a multi-layer perception, and we substitute it with DenseNet-169 to deal with a stronger nonlinearity and image inputs.
The same training, validating, and testing sets as the regression experiment are used. From those sets, 30,000, 10,000, and 20,000 ``pairs'' are randomly created, 
and used for training, validating, and testing RankNet with DenseNet-169.
The cross-entropy loss was used to train RankNet. Again, the standard early stopping rule (no improvement on the validation accuracy for 10 epochs) is used for the termination. \par
Unlike the regression task, the performance of the (bipartite) learning-to-rank task is evaluated the successful ranking rate for the test pairs $\{x_1, x_2\}$. If $x_1$ should be higher ranked than $x_2$ and the learned RankNet correctly evaluates $r(x_1)>r(x_2)$, this is a successful case. The chance rate is 50\%.\par
The successful ranking rates for the 30,000 training pairs and 10,000 test pairs are  60.13\% and 57.19\%, respectively. This is still surprisingly high accuracy because RankNet determines the superiority or inferiority of two given logo images with about 60\% accuracy (i.e., not 50\%). This result also proves that the logo images themselves contain some factor that correlates with the number of followers.

%%%%%%%%%%%%%%%%%%%%%%%%%%%%%%%%%%%%%%%%%%%%%%%%%%%%%%%%%%%
\section{Conclusion}\label{sec:conclusion}
%%%%%%%%%%%%%%%%%%%%%%%%%%%%%%%%%%%%%%%%%%%%%%%%%%%%%%%%%%%
This paper analyzed logo images from various viewpoints. Especially, we focus on three correlations between logo images and their text areas, between the text areas and the number of followers on Twitter, and between the logo images and the number of followers. The first correlation analysis with a state-of-the-art text detector~\cite{baek2019character} revealed that the ratio of logotypes, logo symbols, and mixed symbols is 4\%, 26\%, and 70\%, respectively. In addition, the ratio and the vertical location of the text areas are quantified. The second correlation analysis with DeepCluster~\cite{Caron2018deepcluster} revealed the weak positive correlation between the text area ratio and the number of followers.
The third correlation analysis revealed that deep regression and deep ranking methods could catch some hints for the popularity (i.e., the number of followers) and the relative popularity, respectively, just from logo images.\par
As summarized above, the recent deep learning-based technologies, as well as large public logo image datasets, help 
to analyze complex visual designs, i.e., logo images, in an objective, large-scale, and reproducible manner. Further research attempts, such as the application of explainableAI techniques to deep regression and deep ranking, company-type-wise analysis, font style estimation, etc., will help to understand the experts' knowledge on logo design.

%%%%%%%%%%%%%%%%%%%%%%%%%%%%%%%%%%%%%%%%%%%%%%%%%%%%%%%%%%%
\section{Acknowledgment}\label{sec:Acknowledgment}
%%%%%%%%%%%%%%%%%%%%%%%%%%%%%%%%%%%%%%%%%%%%%%%%%%%%%%%%%%%
This work was supported by JSPS KAKENHI Grant Number JP17H06100.
\bibliographystyle{splncs04}
\bibliography{icdar}




% LNCS Homepage, \url{http://www.springer.com/lncs}. Last accessed 4
% \end{thebibliography}
\end{document}